\documentclass[lettersize,journal]{IEEEtran}
\newcommand{\CLASSINPUTbaselinestretch}{1.35}
\usepackage{cite}
\usepackage{amsmath,amssymb,amsfonts}
\usepackage{algorithm}
\usepackage{algpseudocode}
\usepackage{graphicx}
\usepackage{textcomp}
\usepackage{xcolor}
\usepackage{mathtools}
\usepackage{tikz}
\usetikzlibrary{shapes,arrows}
\usepackage{subcaption}
\usepackage{xcolor,soul}
\usepackage{hyperref}
\usepackage{cases}
\usepackage{siunitx}
\input{mysymbol.sty}
\newtheorem{problem}{\hspace{-10pt}\bf Problem}

\def\BibTeX{{\rm B\kern-.05em{\sc i\kern-.025em b}\kern-.08em
    T\kern-.1667em\lower.7ex\hbox{E}\kern-.125emX}}

\begin{document}

\markboth{IEEE TRANSACTIONS ON SIGNAL PROCESSING (SUBMITTED)}%
         {Zilberstein et al. \MakeLowercase{\textit{et al.}}: \theTitle}

\title{Solving Linear Inverse Problems using Higher-Order Annealed Langevin Diffusion\\
\thanks{This work was partially supported by Nvidia. 
All authors are with the Department of Electrical and Computer Engineering, Rice University, Houston, TX,
USA.
Email: \{nzilberstein, ashu, segarra\}@rice.edu. Preliminary results were published in~\cite{zilberstein2023}.}
}

\author{%
      \IEEEauthorblockN{Nicolas Zilberstein, Ashutosh Sabharwal, Santiago Segarra}  
}

\maketitle
\begin{abstract}
We propose a solution for linear inverse problems based on higher-order Langevin diffusion.
More precisely, we propose pre-conditioned second-order and third-order Langevin dynamics that provably sample from the posterior distribution of our unknown variables of interest while being computationally more efficient than their first-order counterpart and the non-conditioned versions of both dynamics.
Moreover, we prove that both pre-conditioned dynamics are well-defined and have the same unique invariant distributions as the non-conditioned cases.
We also incorporate an annealing procedure that has the double benefit of further accelerating the convergence of the algorithm and allowing us to accommodate the case where the unknown variables are discrete.
Numerical experiments in two different tasks in communications (MIMO symbol detection and channel estimation) and in three tasks for images showcase the generality of our method and illustrate the high performance achieved relative to competing approaches (including learning-based ones) while having comparable or lower computational complexity.
\end{abstract}
%

\section{Introduction}\label{S:intro}

Denoising and super-resolution in images~\cite{image_inverse},  symbol detection and channel estimation in massive MIMO~\cite{Proakis2007}, and graph topology inference~\cite{segarra2017network, dong2016learning}, among others, are all fundamental tasks in our modern world.
Although seemingly unrelated, all these tasks can be formulated as inverse problems. 
Successfully solving these tasks, especially in real-time scenarios, requires the design and implementation of \emph{fast, scalable, and computationally efficient} algorithms.

In many important inverse problems, the recovery of the true unobserved or hidden variables from (noisy) observations is an underdetermined problem, and thus the solution is not unique. 
To overcome this, a key ingredient is the structural or prior information of the signal of interest, which aims to regularize the problem.
In the Bayesian framework, which is the one considered in this paper, the regularization is represented by the prior distribution, and henceforth an inverse problem can be posed as sampling from the posterior distribution of the hidden variables given noisy observations~\cite{idier2013bayesian}.
Then, the solution to the inverse problem might involve computing the value that maximizes the posterior probability distribution (MAP) of the hidden variables given a set of measurements or the conditional expectation, which are point estimations.
However, sampling directly from the posterior distribution is typically intractable.
Therefore, different sampling techniques have been used, most of them based on Markov chain Monte Carlo (MCMC) methods~\cite{MCMCbook}.
In the past few years, deep neural networks have become
ubiquitous in solving inverse problems, showing unprecedented performance~\cite{scarlett2023theoretical}.
In this context, diffusion models based on the discretization of stochastic differential equations~\cite{song2021scorebased} have shown state-of-the-art performance in several inverse problems~\cite{kawar2022denoising, chung2023diffusion}.

In particular, the discretization of the Langevin diffusion, known as \emph{unadjusted Langevin algorithm}, has been used in image denoising~\cite{kawar2021snips, laumont2022bayesian}, MIMO symbol detection~\cite{zilberstein2021, zilberstein2022}, and channel estimation~\cite{arvinte2022}, achieving impressive results.
The unadjusted Langevin algorithm is an iterative technique that enables sampling from the posterior distribution by leveraging the availability of the gradient of the log-posterior distribution, known as the score function. 
In a nutshell, the algorithm starts from a random point and gradually converges to the MAP estimate of the posterior distribution by following the direction of the gradient of the log-posterior distribution.
In addition, an annealing process given by a noise schedule is used to accelerate the convergence of the algorithm. 
Although this method shows outstanding results, its convergence can be slow.
Therefore, in the past few years, higher-order Langevin dynamics have gained interest as they show a better convergence rate. 
In particular,~\cite{pmlr-v75-cheng18a} shows that the \emph{underdamped} (or \emph{second-order}) Langevin dynamic (ULD) has a better convergence rate in the non-asymptotic regime compared to the first-order case.
In essence, a momentum variable is added to the dynamic, which entails smoothing trajectories and thus improves the mixing time, an effect that resembles acceleration in classical gradient descent~\cite{YiAn2021IsTA}.
Moreover, in~\cite{high_order_langevin}, a \emph{third-order} dynamic was proposed, and it was shown that adding a third auxiliary variable entails an even better convergence rate.
This dynamic can be seen as a particular instance of the generalized Langevin equation (GLE), which is a more general model where the evolution of the dynamic depends on the history of the trajectory.

In addition to extending the system by adding auxiliary variables, different pre-conditioning methods have been proposed to adapt the geometry of the search space, and therefore speed up the convergence of the dynamic.
Inspired by Newton's method,~\cite{kawar2021snips} and~\cite{zilberstein2022} use a pre-conditioning matrix in the spectral domain for the overdamped (first-order) case.
From a geometric point of view, this position-dependent pre-conditioning can be interpreted as running the dynamic on a manifold with constant curvature~\cite{girolami2011riemann}.
For the underdamped (second-order) case, pre-conditioning techniques have also been proposed~\cite{Fu2016QuasiNewtonHM, csimcsekli2018asynchronous}. 
Furthermore,~\cite{dockhorn2022scorebased} developed a score-based generative model based on a particular case of the underdamped diffusion, which accelerates the sampling process.
Overall, the dynamics defined in this body of work -- pre-conditioned version of the ULD -- are special cases of the general framework introduced in~\cite{ma2015complete}.
%

\vspace{0.7mm}
\noindent
{\bf Contributions.} Given the better non-asymptotic convergence rate of higher-order Langevin dynamics and the benefits of using a pre-conditioning matrix, we propose a general framework to solve linear inverse problems using an annealed version of both the \emph{underdamped} and \emph{third-order} dynamics that allows the incorporation of a pre-conditioning matrix.
Thus, we seek a framework that strikes a balance between state-of-the-art performance and low-running time complexity.
As far as we know, this is the first time a pre-conditioned third-order Langevin diffusion is proposed.
We show that both modified dynamics are well-defined and have the same unique invariant distribution as the classic dynamics.
The proposed framework allows the inclusion of both statistical and/or learning-based prior information.
Our contributions are three-fold:\\
1) We propose a general framework for solving linear inverse problems based on \emph{annealed higher-order Langevin dynamics} with pre-conditioning matrix, allowing us to reduce the computational complexity compared to the {overdamped} (first-order) case, and we prove that the modified continuous-time dynamics have the same unique invariant distribution as the non-conditioned counterparts\\
2) We analyze the behavior of our proposed dynamics using different discretization schemes based on the splitting operator technique.\\
3) Through numerical experiments, we analyze our proposed framework in several downstream tasks, namely MIMO symbol detection and channel estimation, and Gaussian deblurring, inpainting, and super-resolution.
We demonstrate that both dynamics achieve a better performance than the overdamped case, with lower complexity in terms of running time, and also outperform different baseline methods.

\vspace{0.7mm}
\noindent
{\bf Paper outline.}
In Section~\ref{sec:problem_formulation}, we present the system model and the problem formulation. 
In Section~\ref{sec:CLD}, we briefly introduce the continuous-time ULD and GLE.
In Section~\ref{sec:pre-cond}, we introduce modifications of both continuous-time dynamics to allow the use of a pre-condition matrix, and we show that both dynamics preserve the invariant distribution of their classical counterparts.
Then, we introduce numerical discretization schemes and the algorithms used for sampling from the posterior distribution.
In Section~\ref{sec:numerical_experiments}, we demonstrate the gains in performance and complexity of our proposed method through numerical experimentation.
Finally, Section~\ref{sec:conclusions} wraps up the paper and discusses possible future work.

\vspace{-2mm}
\section{System model and problem formulation}
\label{sec:problem_formulation}

A general noisy linear forward model is described as
\begin{equation}\label{E:mimo_model}
	\bby = \bbH \bbx + \bbz,
\end{equation}
where $\bbH \in \mathbb{R}^{N_r \times N_u}$ is the linear operator (channel matrix in MIMO system, degradation operator in images, etc.), $\bbz \sim \mathcal{N}(\bb0, \sigma_0^2 \bbI_{N_r})$ is a vector of Gaussian noise, $\bbx \in \mathcal{X}^{N_u}$ is the signal of interest, $\mathcal{X}$ is the domain of the signal, which might be a finite or an infinite set, and $\bby \in \mathbb{R}^{N_r}$ is the observed vector. 
Moreover, we assume perfect knowledge of $\sigma_0^2$ and either $\bbH$ or $\bbx$.
Under this configuration, the estimation problem can be {stated} as follows.
\vspace{0.1in}
\begin{problem}\label{P:main} \emph{
	Given perfect knowledge of $\bbH$ (or $\bbx$) and an observed $\bby$ following~\eqref{E:mimo_model}, find an estimate of $\bbx$ (or $\bbH$).}
\end{problem}
\vspace{0.1in}
To facilitate the presentation throughout the paper, we will derive everything for the case when $\bbx$ is unknown; the derivation when $\bbH$ is unknown is equivalent, as we show in the numerical experiments.
Given that $\bbz$ in~\eqref{E:mimo_model} is a random variable, a natural way of solving Problem~\ref{P:main} is to search for the $\bbx$ that maximizes its \emph{posterior} probability given the noisy observations $\bby$.
Hence, Bayes' optimal decision rule can be written as
\begin{align}\label{eq:map}
	\hat{\bbx}_{\mathrm{MAP}} = \argmax_{\bbx \in \mathcal{X}^{N_u}}\,\, p(\bbx|\bby,\bbH)
	= \argmax_{\bbx \in \mathcal{X}^{N_u}}\,\, p_{\bbz}(\bby - \bbH\bbx)p(\bbx).
\end{align}
%
Several schemes have been proposed in the last decades to provide efficient approximate solutions to Problem~\ref{P:main} (see Section~\ref{S:intro}).
Recently, samplers based on Langevin diffusion were proposed for different applications~\cite{kawar2021snips,zilberstein2022,arvinte2022}, achieving high performance.
However, these samplers are demanding in terms of computation and running time.

Hence, we are motivated by the following question: \textit{Can we derive an efficient algorithm to solve inverse problems by sampling from the posterior? 
More specifically, compared to the performance of the overdamped Langevin dynamic, can we reduce its running time but still achieve state-of-the-art performance?}
We achieve this objective by leveraging an annealed version of two different higher-order Langevin dynamics.
In particular, we propose to solve Problem~\ref{P:main} by (approximately) sampling from the posterior distribution in~\eqref{eq:map} using \emph{pre-conditioned} annealed underdamped (second-order) and third-order Langevin dynamics.

\section{Higher-order Langevin dynamics}
\label{sec:CLD}

In Section~\ref{subsec:CTULD}, we introduce the continuous-time ULD, also known as the second-order Langevin dynamic. 
In Section~\ref{subsec:CGLE}, we describe the GLE, which serves as a framework to derive higher-order dynamics.

\subsection{Continuous-time underdamped Langevin dynamics}
\label{subsec:CTULD}

The continuous-time ULD is the Markov process on variables $\bbx_t \in \mathbb{R}^d$ and $\bbv_t \in \mathbb{R}^d$ that solves the stochastic differential equations (SDEs)
\begin{align}\label{eq:ct_underdamped_langevin}
    \text{d}\bbx_t &= \bbM^{-1}\bbv_t \text{d}t,\\\nonumber
    \text{d}\bbv_t &= -\nabla_{\bbx_t} U(\bbx_t)\text{d}t - \gamma \bbv_t\text{d}t + \sqrt{2 \gamma \tau} \bbM^{\frac{1}{2}}\text{d}\bbW_t,
\end{align}
\noindent where $\bbW$ is a standard $d$-dimensional Brownian motion, $U \in \ccalC^2(\mathbb{R}^d)$ is called the potential, $\gamma \in \mathbb{R}^+$ is a friction parameter, $\bbM$ is a mass matrix that controls the coupling between $\bbx_t$ and $\bbv_t$, and $\tau$ is a temperature parameter.
Under mild conditions, it can be shown~\cite{pavliotis_book} that the invariant distribution of the continuous-time process~\eqref{eq:ct_underdamped_langevin} lies in the kernel of the Fokker-Planck equation, and it is given by 
\begin{equation}\label{eq:inv_distr}
    \pi(\bbX) \propto \exp\left[- \tau^{-1}H\left(\bbX\right)\right],
\end{equation}
where $\bbX = [\bbx, \bbv]$ and $H(\bbX) = U(\bbx) + \frac{\bbv^\top \bbM^{-1}\bbv}{2}$ is the Hamiltonian of the system.
Therefore, the state distribution of the process in~\eqref{eq:ct_underdamped_langevin} converges to $\pi(\bbX)$ when $t\rightarrow \infty$.
Given a target distribution $p(\bbx)$ from which we want to generate samples $\bbx \in \mathbb{R}^d$, if we define $U(\bbx) = -\log p(\bbx)$, then~\eqref{eq:ct_underdamped_langevin} defines an MCMC sampler as  $\pi(\bbx) \propto p(\bbx)^{1/\tau}$.
In particular, if $ \tau = 1$, then $\pi(\bbx) \propto p(\bbx)$.
In essence, the dynamic in~\eqref{eq:ct_underdamped_langevin} explores the target distribution by moving in the direction of the gradient of the logarithm of the target density $\nabla_{\bbx}\log p(\bbx)$, known as \emph{score function}.

\vspace{0.5cm}
\noindent {\bf Accelerated mixing time and overdamped (first-order) Langevin as limit dynamic.}
The \emph{overdamped} Langevin dynamic can be obtained as a particular regime of the dynamic in~\eqref{eq:ct_underdamped_langevin}, when the friction parameter $\gamma \xrightarrow{} \infty$~\cite[Section~6.5]{pavliotis_book}.
Therefore, the main difference between the overdamped and the underdamped case is the momentum variable $\bbv_t$.
Through the lens of sampling, the momentum term entails, under some assumptions on the log-probability density $\log p(\bbx)$, an accelerated version of the sampler compared to the overdamped Langevin dynamic.
In particular,~\cite{pmlr-v75-cheng18a} establishes a non-asymptotic convergence rate for a specific discretization scheme of~\eqref{eq:ct_underdamped_langevin} when $\log p(\bbx)$ is {strongly concave} and has Lipschitz continuous gradient (this result was later improved in~\cite{Dalalyan2018OnSF}).
In~\cite{Monmarche2021HighdimensionalMW}, another convergence rate is given by considering a splitting strategy as the discretization scheme; details about discretization schemes are postponed to Section~\ref{subsec:CLDdiscr}.
Overall, these results demonstrate that there is a significant improvement when considering the underdamped case.

\subsection{Generalized Langevin equation}
\label{subsec:CGLE}

The GLE represents a more general stochastic dynamical system represented by the following SDEs
\begin{align}\label{eq:ct_Glangevin}
    \text{d}\bbx_t &= \bbM^{-1}\bbv_t \text{d}t,\\
    \text{d}\bbv_t &= -\nabla_{\bbx_t} U(\bbx_t) - \bigg[ \int_0^t \bbK(t-s)\bbM^{-1}\bbv_s\text{d}s\bigg]\text{d}t + \bbeta_t\text{d}t,\nonumber
\end{align}
where $\bbK(t)$ is a memory kernel defining a drag force (or linear dissipation term) in the form of a convolution with the momentum. 
Therefore, the dissipation term depends on the history of the position.
The stochastic force $\bbeta_t$ is related to the dissipation term through the fluctuation-dissipation theorem~\cite{pavliotis_book}.
In particular, $\bbeta_t$ is defined as a stationary Gaussian process with zero mean and auto-covariance function given by
\begin{align}\label{eq:fluct-dissi}
\E{\bbeta(s + t)\bbeta^\top(s)} =\tau\bbK(t),
\end{align}
where $\tau$ is the temperature of the system.
The GLE is a {non-Markovian dynamical model, in the sense that the evolution of the system depends on the history of the trajectory through the convolution integral in the momentum equation.}
Therefore, from a numerical point of view, its implementation is computationally demanding as it requires the storage of the momentum history -- whose numerical evaluation at each time step can be non-trivial -- as well as the generation of a stochastic force with non-trivial correlations.
Under certain conditions on the memory kernel $\bbK(t)$, the GLE is ergodic with unique invariant distribution given by the Gibbs distribution~\eqref{eq:inv_distr}. 
Thus, we can also generate samples from a target distribution $p(\bbx)$ by defining $U(\bbx) = -\log p(\bbx)$.

Although the inclusion of the memory kernel enhances the sampling properties~\cite{ceriotti2010novel}, it makes this formulation analytically intractable.
Therefore, we can circumvent this issue by working with a particular choice of the memory kernel, which allows us to reformulate the dynamic in a Markovian representation by adding a set of auxiliary variables.

\vspace{0.5cm}
\noindent \textbf{Extended-variable Markovian representation.} To rewrite the SDEs in~\eqref{eq:ct_Glangevin} in a Markovian representation, we define the memory kernel as $\bbK(t)= K(t)\bbI$, where $K(t)$ is a scalar-valued function.
Under this representation of the kernel, we define the function $K(t)$ as a Prony series
\begin{align}\label{eq:kernel}
    K(t) = \sum_{j=1}^{j_{\max}}\lambda_j^2\exp{\left(-\alpha_j t\right)},
\end{align}
where there are $j_{\max}$ modes, and each $\lambda_j$ and $\alpha_j$ are constant parameters. 
With this choice of $K(t)$, we can represent the GLE~\eqref{eq:ct_Glangevin} as a Markovian system by introducing $j_{\max}$ additional variables. 
From now on, we consider only one Prony mode, i.e., $j_{\max} = 1$, so that we drop the subindex $j$; the dynamic that we derive can be extended easily to the case $j_{\max} > 1$~\cite{Leimkuhler}.

Formally, we define the extended variable $\bbxi$ associated with the Prony mode as the dissipation term given by the convolution operation in~\eqref{eq:ct_Glangevin} divided by $\lambda$
\begin{align}\label{eq:extended_variable}
   \bbxi_t = -\frac{1}{\lambda} \int_0^t \lambda^2\exp{\left(-\alpha(t-s)\right)}\bbM^{-1}\bbv_s \text{d}s.
\end{align}
%
Then, we can rewrite~\eqref{eq:ct_Glangevin} as
\begin{align}\label{eq:ct_Glangevin_prony}
    \text{d}\bbx_t &= \bbM^{-1}\bbv_t \text{d}t\\
    \text{d}\bbv_t &= -\nabla_{\bbx_t} U(\bbx_t)\text{d}t + \lambda\bbxi_t\text{d}t + \bbeta_t\text{d}t,\nonumber
\end{align}
Now, we redefine the drag force and the stochastic force to obtain a Markovian system.  

First, using the Laplace transform, it can be seen that the variable $\bbxi_t$ defined in~\eqref{eq:extended_variable} is the solution of the ordinary differential equation given by
\begin{equation}\label{eq:drag_force}
    \text{d}\bbxi_t = -\alpha\bbxi_t\text{d}t - \lambda\bbM^{-1}\bbv_t\text{d}t.
\end{equation}
Second, we construct a random force for the Prony mode by noticing that the random force has to satisfy the fluctuation-dissipation theorem~\cite{kubo1966fluctuation}, so it can be viewed as the solution of a linear SDE 
\begin{equation}\label{eq:stoch_force}
    \text{d}\tilde{\bbeta}_t = -\alpha \tilde{\bbeta}_t\text{d}t + \sqrt{2 \tau \alpha}\text{d}\bbW_t.
\end{equation}
Thus, the random force is defined as the solution of an Ornstein-Uhlenbeck process induced by the Prony series.
Moreover, we need to multiply the solution by $\lambda$ to satisfy the fluctuation-dissipation theorem, i.e., $\bbeta_t = \lambda \tilde{\bbeta}_t$.
Hence, the process $\bbeta_t$ has zero mean and time correlation function given by $\E{\bbeta(s+t)\bbeta(s)^\top} = \tau \lambda^2 \exp\left(-\alpha t\right)\bbI$.

We can combine both results~\eqref{eq:drag_force} and~\eqref{eq:stoch_force} and define a new variable $\bbz_t = \bbxi_t + \tilde{\bbeta}_t$, to reformulate the GLE as a Markovian system
\begin{align}\label{eq:ct_G_markov_langevin_3rdorder}
    \text{d}\bbx_t &= \bbM^{-1}\bbv_t \text{d}t,\\
    \nonumber\text{d}\bbv_t &= -\nabla_{\bbx_t} U(\bbx_t)\text{d}t + \lambda\bbz_t\text{d}t,\\
    \nonumber\text{d}\bbz_{t} &= - \lambda\bbM^{-1}\bbv_t\text{d}t - \alpha\bbz_{t}\text{d}t + \sqrt{2 \tau \alpha} \, \text{d}\bbW_{t}.
\end{align}
Under this representation of the GLE, we can define a Fokker-Planck operator associated with the dynamic and, consequently, an invariant distribution of the dynamic.
In comparing the GLE with the (second-order) ULD, we first notice that the latter can be seen as a particular case of the former when $K(t) = \gamma \delta(t)$~\cite{ottobre2011asymptotic}.
Moreover, in~\cite{high_order_langevin}, it was shown that the third-order case generates smoother trajectories than the corresponding ones under the ULD, which enhances the sampling properties and exhibits an improvement in the convergence rate in comparison to the overdamped and underdamped cases. 
Also,~\cite{chak2020generalised} introduced an annealing scheme that shows better performance and higher exploration of the state space compared to the underdamped and overdamped cases.
\section{Accelerating higher-order Langevin dynamics}
\label{sec:pre-cond}

In Section~\ref{subsec:pre_cond}, we define two modified continuous-time Markovian Langevin dynamics, one associated with the second-order and the other one with the third-order dynamic.
We show that the two processes have a unique invariant distribution given by~\eqref{eq:inv_distr}.
Then, in Section~\ref{subsec:CLDdiscr}, we explain the numerical implementation of these dynamics via discretization schemes based on the splitting method.
Lastly, in Section~\ref{subsec:posterior}, we derive our posterior sampling algorithms, where we present different possible choices of the score function of the posterior distribution and the pre-conditioning matrix involved in the algorithms.

\vspace{-0.3cm}   
\subsection{Modified continuous-time higher-order Langevin dynamics}
\label{subsec:pre_cond}

This section aims to construct continuous-time diffusion processes based on the dynamics introduced in Section~\ref{sec:CLD} that leverage the geometry of the space to accelerate the convergence. 
We achieve this acceleration by incorporating carefully crafted pre-conditioning
denoted by $\bbC_t \in \mathbb{R}^{d\times d}$.
We assume the following for $\bbC_t$ 
\vspace{0.2cm}
\assumption{\label{Ass1}The matrix $\bbC_t \in \mathbb{R}^{d\times d}$ is symmetric positive definite and has uniformly bounded matrix norm, i.e., $||\bbC_t|| < M$ for all $t$.}
\vspace{0.2cm}
%

Before introducing the dynamics, we assume also the following for the potential function.
\vspace{0.2cm}
\assumption{\label{Ass2}The gradient of the potential is Lipschitz continuous, i.e., 
$||\nabla_{\bbx} U(\bbx) - \nabla_{\bbx} U(\bbx)|| \leq L ||\bbx - \bbx'||, \quad \forall \; \bbx, \bbx' \in \mathbb{R}^d$.}
\vspace{0.2cm}

This assumption is a necessary condition to guarantee the uniqueness of the invariant distribution of our proposed pre-conditioned dynamics.

\vspace{0.1cm}
\noindent \textbf{Pre-conditioned ULD.} 
As defined in Section~\ref{subsec:CTULD}, this dynamic has two variables: position and momentum.
Thus, we have $\bbX_t = [\bbx_t, \bbv_t]$ and 
\begin{equation}\label{eq:hamiltonian_ULD}
    H(\bbX_t) = U(\bbx_t) + \frac{1}{2} \bbv_t^\top \bbM^{-1}\bbv_t,
\end{equation}
and the gradient $\nabla H (\bbX_t) = [\nabla_{\bbx_t} U (\bbx_t), \bbM^{-1}\bbv_t]$.
Moreover, we consider a symmetric positive definite matrix $\bbC_t \in \mathbb{R}^{d\times d}$.
Then, the pre-conditioned second-order dynamic is defined as follows
\begin{align}\label{eq:ct_underdamped_langevin_precond}
    \text{d}\bbx_t &= \bbC_t\bbM^{-1}\bbv_t \text{d}t,\\
    \nonumber \text{d}\bbv_t &= -\bbC_t\nabla_{\bbx_t} U(\bbx_t) - \gamma \bbv_t\text{d}t + \sqrt{2 \tau \gamma} \bbM^{\frac{1}{2}}\text{d}\bbW_t,
\end{align}
The following proposition guarantees that the invariant distribution is still~\eqref{eq:inv_distr}.
\vspace{0.2cm}
\proposition{\label{prop1}Given Assumptions~\ref{Ass1} and~\ref{Ass2} and the Hamiltonian defined in~\eqref{eq:hamiltonian_ULD}, the distribution~\eqref{eq:inv_distr} is the unique invariant distribution of the dynamics~\eqref{eq:ct_underdamped_langevin_precond}.}
\vspace{0.2cm}

The proof is deferred to Appendix~\ref{A1}. 
This proposition generalizes existing results~\cite{leimkuhler2018ensemble} and~\cite{csimcsekli2018asynchronous} by accounting for the effect of a mass matrix $\bbM$.

\vspace{0.1cm}
\noindent \textbf{Pre-conditioned third-order LD.} Following a similar recipe to the ULD case, we can derive a pre-conditioned version of the dynamical system defined in~\eqref{eq:ct_G_markov_langevin_3rdorder}.
First, the variables involved in this dynamic are $\bbX_t = [ \bbx_t, \bbv_t, \bbz_t]$, so that%
\begin{equation}\label{eq:hamiltonian_3rd}
    H(\bbX_t) = U(\bbx_t) + \frac{1}{2} \bbv_t^\top \bbM^{-1}\bbv_t + \frac{1}{2} \bbz_t^\top \bbM^{-1}\bbz_t,
\end{equation}
and the gradient is $\nabla H (\bbX_t) = [\nabla_{\bbx_t} U (\bbx_t), \bbM^{-1}\bbv_t, \bbM^{-1}\bbz_t]$.
Then, the pre-conditioned third-order dynamic is defined as follows\footnote{The square root of $\bbM$ is multiplying the stochastic term, while in~\eqref{eq:ct_underdamped_langevin} the inverse of $\bbM$ is multiplying the velocity variable. 
Both representations are equivalent, where the difference comes from a re-scaling of the kernel in~\eqref{eq:kernel} with $\bbM$. 
Then, the fluctuation-dissipation theorem~\eqref{eq:fluct-dissi} has to hold, so the stochastic term is multiplied by $\bbM^{1/2}$.}
\begin{align}\label{eq:ct_G_markov_langevin_precondition}
    \text{d}\bbx_t &= \bbC_t\bbM^{-1}\bbv_t \text{d}t,\\
    \nonumber\text{d}\bbv_t &= -\bbC_t\nabla_{\bbx_t} U(\bbx_t)\text{d}t + \lambda\bbz_t\text{d}t,\\
    \nonumber\text{d}\bbz_{t} &= - \lambda\bbv_t\text{d}t - \alpha\bbz_{t}\text{d}t  + \sqrt{2\tau \alpha}\bbM^{1/2}\text{d}\bbW_{t},
\end{align}
%
Again, the following proposition guarantees that the invariant distribution is~\eqref{eq:inv_distr}.
\vspace{0.2cm}
\proposition{\label{prop2}Given Assumptions~\ref{Ass1} and~\ref{Ass2} and the Hamiltonian defined in~\eqref{eq:hamiltonian_3rd}, the distribution~\eqref{eq:inv_distr} is the unique invariant distribution of the dynamics~\eqref{eq:ct_G_markov_langevin_precondition}.}
\vspace{0.2cm}

The proof is deferred to Appendix~\ref{A2}. 
\vspace{0.2cm}

Considering a pre-conditioning matrix is key in real-time applications, as it accelerates the convergence of the dynamic while yielding good performance.
Recently, pre-conditioned \emph{overdamped} Langevin dynamics have been proposed for denoising tasks in image processing~\cite{kawar2021snips} and MIMO detection in communications~\cite{zilberstein2022}, where the negative inverse of the Hessian matrix of the score-posterior is used as a pre-conditioning matrix.
Also, in~\cite{ma2022accelerating,zhang2023preconditioned}, a general recipe for designing the pre-conditioning matrix was proposed for the overdamped Langevin dynamics.
Intuitively, the matrix takes into account the local geometry by adding second-order information to the exploration of the search space.
This translates to a better convergence rate of the dynamic in all directions.
Here we have extended this analysis, enabling the use of pre-conditioned higher-order Langevin dynamics for the solution of inverse problems.


\subsection{Numerical discretization}
\label{subsec:CLDdiscr}

In order to define a sampling algorithm based on the continuous-time dynamics introduced in Section~\ref{subsec:pre_cond}, we need to discretize the equation.
In this work, we rely on splitting methods for the discretization of~\eqref{eq:ct_underdamped_langevin_precond} and~\eqref{eq:ct_G_markov_langevin_precondition}. 
In a nutshell, given the operator that describes the time evolution of the state $\bbx_t$, the idea is to split the operator into tractable sub-operators and then compose them to approximate the full operator. 
Formally, given an initial state $\bbX_0$, the solution of the SDE~\eqref{eq:ct_underdamped_langevin} can be constructed as $\bbX_t = \exp{(t\ccalL_{\mathrm{ULD}})} \bbX_0$, where $\exp(t\ccalL_{\mathrm{ULD}})$ is the operator that defines the propagation of the state and $\ccalL_{\mathrm{ULD}}$ is the (infinitesimal) generator of the Markov process~\cite{pavliotis_book}. 
Then, we split the operator $\ccalL_{\mathrm{ULD}}$ into sub-operators~\cite{Leimkuhler}, and rewrite the dynamic as the composition of each sub-operation.
Since we focus on sampling throughout this work, we work directly with $U(\bbx_t) = -\log p(\bbx_t)$ in deriving the numerical discretization schemes.

\vspace{0.2cm}
\noindent \textbf{Numerical discretization of ULD.} 
For the ULD~\eqref{eq:ct_underdamped_langevin_precond}, we rewrite the dynamic as

\begin{align}
    \hspace{-1mm}\text{d}
    \begin{bmatrix}\bbx_t \\ \bbv_t 
    \end{bmatrix} = 
    \underbrace{
    \begin{bmatrix}\bbC_t\bbM^{-1}\bbv_t\\ \bb0 
    \end{bmatrix}\text{d}t}_{\text{A}} + \underbrace{
    \begin{bmatrix}\bb0 \\  \bbC_t\nabla_{\bbx_t}\log p(\bbx_t)
    \end{bmatrix}\text{d}t}_{\text{B}} +& \\
    \nonumber &\hspace{-3cm}\underbrace{
    \begin{bmatrix} \bb0 \\ - \gamma \bbv_t\text{d}t + \sqrt{2 \gamma \tau} \bbM^{\frac{1}{2}}\text{d}\bbW\end{bmatrix}}_{\text{O}}. \nonumber
\end{align}

\noindent The terms labeled $\text{A}$ and $\text{B}$ refer to the Hamiltonian components, which can be solved using any deterministic numerical integrator, and the termed labeled O refers to the Ornstein-Uhlenbeck process, {which has a closed-form expression when integrating it in an interval $[k\epsilon, (k+1)\epsilon)$}.
Given this splitting strategy, we can derive a family of schemes. 
The simpler one, denoted by ABO, has the following state propagation in an interval of length $\epsilon$
\begin{equation}
    \exp(\epsilon\ccalL_{\mathrm{ULD}}) = \exp(\epsilon\ccalL_{O})\exp(\epsilon\ccalL_{B})\exp(\epsilon\ccalL_{A}).
\end{equation}
Therefore, given a step size $\epsilon$, the discretization scheme is given by
\begin{align}\label{eq:discrete_langevin}
\bbx_{k+1} &= \bbx_{k} + \epsilon \bbC_{\epsilon}\bbM^{-1}\bbv_{k}, \\
\nonumber \bbv_{k+1/2} &= \bbv_{k} + \epsilon\bbC_{\epsilon}\nabla_{\bbx_{k+1}}\log p(\bbx_{k+1}), \\
\nonumber \bbv_{k+1} &= \exp(-\gamma \epsilon)\bbv_{k+1/2} + \sqrt{\tau(1-\exp(-2\gamma \epsilon))}\bbM^{\frac{1}{2}}\bbw_k,
\end{align}

\noindent \normalsize where $\bbw_k \sim \ccalN(0, \bbI)$.
This particular scheme corresponds to the adjoint symplectic Euler scheme to solve the Newtonian part of the Langevin dynamics SDE, followed by an exact Ornstein-Uhlenbeck solution.
In general, we can construct different methods by interleaving and propagating each sub-operator $k$ times, i.e., we propagate each sub-operator with a step-size $\frac{\epsilon}{k}$; propagating for a total time of $\epsilon$ is a requirement to have a consistent method~\cite{Leimkuhler}.
As a result, the convention is that if each letter associated with each sub-operator appears $k$ times, then it is updated with step-size $\frac{\epsilon}{k}$.
For example, the symmetric Langevin velocity-Verlet, denoted  by BAOAB, is given by the following state propagation
\begin{align}
      \exp(\epsilon\ccalL_{\mathrm{ULD}}) = \exp  \bigg(\frac{\epsilon}{2}\ccalL_{B}\bigg)&\exp\bigg(\frac{\epsilon}{2}\ccalL_{A}\bigg)\exp(\epsilon\ccalL_{O}) \\
      \nonumber&\exp\bigg(\frac{\epsilon}{2}\ccalL_{A}\bigg)\exp\bigg(\frac{\epsilon}{2}\ccalL_{B}\bigg),
\end{align}
where $k=2$ for the components A and B. 
We do not write the discretization obtained from this scheme as the procedure to obtain the equations is equivalent to the ABO. 

The numerical integration scheme used to discretize~\eqref{eq:ct_underdamped_langevin} largely determines the performance of the algorithm~\cite{jain2022journey}, as we will show in Section~\ref{subsec:channel_estimation}.
Although there are simpler schemes like the \emph{Euler-Maruyama discretization}, which is a first-order integrator, we use the splitting method in~\eqref{eq:discrete_langevin} as it shows better empirical performance.

\vspace{0.5cm}
\noindent \textbf{Numerical discretization for third-order LD.} 
Following a similar strategy to the ULD, we rewrite the dynamic in~\eqref{eq:ct_G_markov_langevin_precondition} as
\begin{align}
    \hspace{-1mm}\text{d}\begin{bmatrix}\bbx_t \\ 
    \bbv_t \\ 
    \bbz_{t} 
    \end{bmatrix} = 
    \underbrace{\begin{bmatrix}\bbC_t\bbM^{-1}\bbv_t \\
    \bb0 \\ \bb0 
    \end{bmatrix}\text{d}t}_{\text{A}} + &
    \underbrace{\begin{bmatrix} \bb0 \\ 
    \bbC_t\nabla_{x_t}\log p(\bbx_t) \\ 
    \bb0 \end{bmatrix}\text{d}t}_{\text{B}} +  \\ &
    \hspace{-3.1cm}
    \underbrace{\begin{bmatrix} \bb0 \\ 
    \lambda \bbz_{t} \\ 
    \bb0 \end{bmatrix}\text{d}t}_{\text{C}} +
    \underbrace{\begin{bmatrix} \bb0 \\ 
    \bb0 \\ - \lambda\bbv_t\text{d}t -\alpha\bbz_{t}\text{d}t + \sqrt{2\tau \alpha}\bbM^{1/2}\text{d}\bbW_{t} \end{bmatrix}}_{\text{O}}. \nonumber
\end{align}
In this case, we consider also two possible discretization methods.
For the first one, we combine the sub-operators B and C and discretized this combination with a velocity Verlet numerical integrator~\cite{baczewski2013numerical}.
We denote this method by (BC)OA(BC) and is given by
\begin{align}
    \bbv_{k+1/2} &= \bbv_{k} + \frac{\epsilon}{2} \bbC_{\epsilon}\nabla_{x_k}\log p(\bbx_k) + \frac{\epsilon}{2}\lambda\bbz_{k}, \label{eq:discr_scheme_1}\\
    \bbx_{k+1} &= \bbx_{k} + \epsilon \bbC_{\epsilon}\bbM^{-1}\bbv_{k+1/2}, \label{eq:discr_scheme_2}\\
    \bbz_{k+1} &= \theta \bbz_{k} - (1 - \theta)\frac{\lambda}{\alpha}\bbv_{k+1/2} + \kappa\sqrt{\tau} \bbM^{1/2}\bbw_{k}, \label{eq:discr_scheme_3}\\
    \bbv_{k+1} &= \bbv_{k+1/2} + \frac{\epsilon}{2} \bbC_{\epsilon}\nabla_{x_{k+1}}\log p(\bbx_{k+1}) + \frac{\epsilon}{2}\lambda \bbz_{k+1}, \label{eq:discr_scheme_4}
\end{align}
where $\theta = \exp{\left(-\epsilon\alpha \right)}$ and $\kappa = \sqrt{1-\theta^2}$~\cite{baczewski2013numerical}.
Notice that we need to propagate each sub-operator for a total time of $\epsilon$.
Hence, each application of the $\ccalL_{BC}$ operator -- \eqref{eq:discr_scheme_1} and~\eqref{eq:discr_scheme_4} -- has a step size of $\epsilon/2$.
The other method that we use in Section~\ref{subsec:channel_estimation} is denoted by BACOCAB, following the convention that we introduced previously.


\subsection{Sampling from the posterior via discrete annealed pre-conditioned higher-order Langevin}
\label{subsec:posterior}

Recall that our goal is to solve Problem~\ref{P:main} by sampling (approximately) from the posterior defined in~\eqref{eq:map} using the discrete dynamics introduced in Section~\ref{subsec:CLDdiscr}.
However, notice that they do not apply directly to Problem~\ref{P:main} as \emph{we do not seek to sample from $p(\bbx)$, but rather from the posterior $p(\bbx|\bby, \bbH)$}.
Thus, in our case, the score is given by $\nabla_{\bbx}\log p(\bbx| \bby, \bbH)$, which can be written after applying Bayes' rule as 
\begin{equation}\label{E:score_function}
\nabla_{\bbx}\log p(\bbx|\bby, \bbH) = \nabla_{\bbx}\log p(\bby|\bbx,\bbH) + \nabla_{\bbx}\log p(\bbx),
\end{equation}
where the term $\nabla_{\bbx}\log p(\bby|\bbx,\bbH)$ corresponds to the guidance (or likelihood) score function and $\nabla_{\bbx}\log p(\bbx)$ to the score function of the prior.
In addition, in this work, we leverage an annealing process, which accelerates the convergence of the dynamic~\cite{ermon2019}.
Therefore, in this section, we present our algorithms based on the annealed version of both the pre-conditioned underdamped and third-order Langevin dynamics.
Below, we give details on different strategies for defining both the guidance score function and the score of the prior for an annealed version of~\eqref{E:score_function}.

\vspace{2mm}
\noindent \textbf{Annealing process.}
We consider an annealed version of our dynamic, where random noise of decreasing variance is added to our state $\bbx$.
This procedure helps accelerate convergence and allows us to accommodate discrete states $\bbx$~\cite{zilberstein2023}.
First, we define a sequence of noise levels $\{\sigma_l\}_{l=1}^{L+1}$ such that $\sigma_1 > \sigma_2 > \cdots > \sigma_L > \sigma_{L+1} = 0$. 
Then, at each level, we define a perturbed version of the true signal $\bbx$ as
\begin{equation}\label{eq:pert_symbs}
    \tilde{\bbx}_{l} = \bbx + \bbn_{l},
\end{equation}
where $\bbn_{l} \sim \mathcal{N}(0, \sigma_l^2\bbI)$.
Given the perturbed signal in~\eqref{eq:pert_symbs}, the forward model in~\eqref{E:mimo_model} can be rewritten as
\begin{align}\label{eq:forwardmodel_noise}
	\bby &= \bbH\tilde{\bbx}_l + (\bbz - \bbH\bbn_l).
\end{align}
Therefore, the annealing process entails a posterior distribution for each noise level $l$ given by
[cf.~\eqref{E:score_function}]
\begin{equation}\label{E:score_function_spectral}
\nabla_{\tilde{\bbx}_l}\!\log p(\tilde{\bbx}_l| \bby, \bbH) = \nabla_{\tilde{\bbx}_l} \log p(\bby|\tilde{\bbx}_l, \bbH) + \nabla_{\tilde{\bbx}_l}\log p(\tilde{\bbx}_l).
\end{equation}
Thus, we have to specify both constituent terms in this score function. 
We now provide details of different strategies for computing each term.
\vspace{1.5mm}

\noindent \emph{i) Score of the guidance term (or likelihood):} In the new forward model in~\eqref{eq:forwardmodel_noise}, the likelihood is given by $p(\bby|\tilde{\bbx}_l, \bbH) = p(\bbz - \bbH {\bbn}_l|\tilde{\bbx}_l)$, which is not Gaussian: although $p(\bbn_l)$ is a Gaussian distribution, after conditioning on $\tilde{\bbx}_l$ the conditional distribution $p(\bbn_l|\tilde{\bbx}_l)$ is no longer Gaussian. 
Therefore, to circumvent this, we consider \textit{two different approximations}, one for each system model considered in Section~\ref{sec:numerical_experiments}.
For the MIMO channel estimation problem, we rely on the approximation introduced in~\cite{arvinte2022, jalal2021}, where we add heuristically a sequence of annealing terms $\{\gamma_l\}_{l=0}^L$ such that $\nabla_{\tilde{\bbx}_l} \log p(\bby|\tilde{\bbx}_l, \bbH) \approx \frac{\bbH^{\text{H}}(\bby - \bbH\tilde{\bbx}_l)}{\sigma_0^2 + \gamma_l^2}$, with $\bbH^{\text{H}}$ denoting the conjugate transpose of $\bbH$.
The second approximation, which we use for the MIMO symbol detection problem, defines the annealing process in the singular value decomposition (SVD) domain.
This was considered in~\cite{zilberstein2022, kawar2021snips}, and it entails a closed-form expression.


\vspace{1.5mm}
\noindent \emph{ii) Score of the prior:} The score function can be related to the MMSE denoiser through Tweedie's identity~\cite{TweedieIdent} as follows
\begin{equation}\label{eq:prior}
	\nabla_{\tilde{\bbx}_l}\log p(\tilde{\bbx}_l) = \frac{\mathbb{E}_{\sigma_l}[\bbx|\tilde{\bbx}_l] - \tilde{\bbx}_l}{\sigma_l^2}.
\end{equation}
When we have access to the conditional expectation, we can compute this score in closed-form~\cite{zilberstein2022}.
Otherwise, we can parameterize it using a neural network; this parameterization is known as a \emph{score network}~\cite{ermon2019}.
Then, we can train it via denoising score matching~\cite{vincent2011connection}.

\begin{algorithm}[t]
	\caption{Pre-conditioned annealed ULD with ABO discretization}\label{alg_uld}
	\begin{algorithmic}
		\Require $T, L, \{\sigma_l, \bbM_l, \bbC_l, \epsilon_l\}_{l=1}^L, \sigma_0, \bbH, \bby, \tau$
		\State Initialize $\tilde{\bbx}_{t=0,l=1}, \bbv_{t=0,l=1}$ randomly
		\For{$l = 1\; \text{to}\;  L$}
		\For{$k = 0\; \text{to}\; T-1$}
			\State Draw $\bbw_k \sim \ccalN(0, \bbI)$

            \State Compute $\nabla_{\tilde{\bbx}_{k,l}}\!\log p(\tilde{\bbx}_{k,l}| \bby, \bbH)$

            \State $\tilde{\bbx}_{k+1, l} = \tilde{\bbx}_{k,l} + \epsilon_l\bbC_l \bbM_l^{-1}\bbv_{k, l}$
            \State $\bbv_{k+1/2, l} = \bbv_{k, l} + \epsilon_l\bbC_l \nabla_{\tilde{\bbx}_{k+1,l}}\!\log p(\tilde{\bbx}_{k+1,l}| \bby, \bbH)$
            \State $\bbv_{k+1, l} \! = \! e^{-\gamma \epsilon_l}\bbv_{k+1/2, l} + \sqrt{\tau (1-e^{-2\gamma\epsilon_l})}\bbM_l^{\frac{1}{2}}\bbw_k$
		\EndFor
		\State $\tilde{\bbx}_{0, l+1} = \tilde{\bbx}_{T, l}$, $\bbv_{0, l+1} = \bbv_{T, l}$
		\EndFor \\
	\Return $\tilde{\bbx}_{T, L}$
	\end{algorithmic}
\end{algorithm}

\begin{algorithm}[t]
	\caption{Pre-conditioned annealed third-order LD with (BC)OA(BC) discretization}\label{alg_3ld}
	\begin{algorithmic}
		\Require $T, L, \{\sigma_l, \bbM_l, \bbC_l, \epsilon_l\}_{l=1}^L, \sigma_0, \bbH, \bby, \tau, \alpha, \lambda$
		\State Initialize $\tilde{\bbx}_{t=0,l=1}, \tilde{\bbv}_{t=0,l=1}, \tilde{\bbz}_{t=0,l=1}$ randomly
		\For{$l = 1\; \text{to}\;  L$}
		\State $\theta_l = \exp{\left(-\epsilon_l \alpha \right)}$
        \State $\kappa_l = \sqrt{1-\theta_l^2}$
		\For{$k = 0\; \text{to}\; T-1$}
			\State Draw $\bbw_k \sim \ccalN(0, \bbI)$

            \State Compute $\nabla_{\tilde{\bbx}_{k,l}}\!\log p(\tilde{\bbx}_{k,l}| \bby, \bbH)$

            \State 
            $\bbv_{k+1/2,l} =  \bbv_{k,l} + \frac{\epsilon_l}{2}  \bbC_l \nabla_{\tilde{\bbx}_{k,l}}\!\log p(\tilde{\bbx}_{k,l}| \bby, \bbH)$
            \State \hspace{1.5cm}$+ 
 \frac{\epsilon_l}{2} \lambda \bbz_{k,l}$
            
            \State $\tilde{\bbx}_{k+1, l} = \tilde{\bbx}_{k, l} + \epsilon_l \bbC_l\bbM_l^{-1}\bbv_{k+1/2, l}$
            
            \State $\bbz_{k+1, l} = \theta_l \bbz_{k, l} - (1 - \theta_l)\frac{\lambda}{\alpha}\bbv_{k+1/2, l} + \kappa_l\sqrt{\tau} \bbM_l^{1/2} \bbw_{k}$
            
            \State $\bbv_{k+1, l} = \bbv_{k+1/2, l} + \frac{\epsilon_l}{2}  \bbC_l  \nabla_{\tilde{\bbx}_{k,l}}\!\log p(\tilde{\bbx}_{k,l}| \bby, \bbH)$
            \State \hspace{1.5cm}$+ \frac{\epsilon_l}{2}\lambda \bbz_{k+1, l}$
		\EndFor
		\State $\tilde{\bbx}_{0, l+1} = \tilde{\bbx}_{T, l}$, $\bbv_{0, l+1} = \bbv_{T, l}$, $\bbz_{0, l+1} = \bbz_{T, l}$
		\EndFor \\
	\Return $\tilde{\bbx}_{T, L}$
	\end{algorithmic}
\end{algorithm}
%
%
%
%
\vspace{2mm}
\noindent {\bf Convergence with score-based priors.} 
Using score-based priors might affect the convergence of the method. 
This was studied in detail in~\cite{laumont2022bayesian} for sampling and in~\cite{laumont2023maximum} for optimization for the case of overdamped Langevin dynamics.
Moreover, in~\cite{sun2023provable}, a convergence analysis for the annealed version of the overdamped case was proposed.
Although a detailed analysis of this situation for higher-order Langevin dynamics is out of the scope of this paper, our experiments in Section~\ref{sec:numerical_experiments} indicate that these results might be generalizable to higher orders.

\vspace{2mm}
\noindent {\bf Algorithm.} 
The algorithms to generate samples $\hat{\bbx}$ from the (approximate) posterior $p(\bbx|\bby,\bbH)$ using ULD and the third-order LD are shown in Algorithm~\ref{alg_uld} and Algorithm~\ref{alg_3ld}, respectively.
Notice that our methods allow the inclusion of a pre-conditioning matrix $\bbC_l$ for each level of noise $l$, by considering a piecewise constant discretization of $\bbC_t$ in~\eqref{eq:ct_underdamped_langevin_precond} and~\eqref{eq:ct_G_markov_langevin_precondition}.

Intuitively, the algorithm works as follows. First, we initialize $\bbx_0, \bbv_0$ -- and $\bbz_0$ for the third-order LD -- randomly.
Then, at each level $l$ of the $L$ noise levels we run $T$ iterations of the ULD or third-order LD, following the direction of the score function of the log-posterior density of the perturbed signal $\nabla_{\tilde{\bbx}_l}\!\log p(\tilde{\bbx}_l| \bby, \bbH)$.
We start a high noise level, so the contribution of the score of the prior is negligible and can be ignored, meaning that the Langevin dynamic considers the observation as a denoised estimation.
Then, we gradually decreases its value until $\tilde{\bbx} \approx \bbx$; in the low noise levels, the idea is to refine the estimation by considering the prior information combined with the guidance term.
Apart from enabling the approximation of the score function of the prior distribution, the annealing process also improves the mixing time of the Langevin dynamic~\cite{ermon2019}. 
Notice that the auxiliary variables are defined at each noise level.
Finally, similar algorithms can be consider for other discretization schemes, like the ones we used in Section~\ref{subsec:channel_estimation}.
We omit them here to avoid redundancy.

Regarding the computational complexity, it strongly depends on the expression of~\eqref{E:score_function_spectral}.
Thus, we define the computational complexity for each system in Section~\ref{sec:numerical_experiments}.

\section{Numerical methods}\label{sec:numerical_experiments}

We start by applying our framework to the symbol detection problem in a massive MIMO system.
In particular, we compare the symbol error rate (SER) performance of both algorithms introduced in Section~\ref{subsec:posterior} in comparison with the annealed overdamped case.
Then, we compare our methods with both classical and learning-based baseline detectors and, finally, we analyze the complexity in terms of running time. 
Second, in Section~\ref{subsec:channel_estimation}, we consider the problem of channel estimation.
We first study the normalized mean squared error (NMSE) performance of our methods considering different discretization schemes, and then their performance compared to the overdamped case.
Finally, we compare with baseline methods and show that the third-order dynamic outperforms all the other methods.
\footnote{Code to replicate the numerical experiments can be found
at \url{https://github.com/nzilberstein/higher-order-langevin}.}

\subsection{Massive MIMO detection}
\label{subsec:detection}

In this section, we analyze the proposed methods for the problem of massive MIMO symbol detection.
This problem can be cast as~\eqref{E:mimo_model}, where $N_u$ are the number of single-antenna transmitters or users and $N_r$ the antennas at the base station.
In this context, $\bar{\bbH} \in \mathbb{C}^{N_r \times N_u}$ is the channel matrix, $\bar{\bbz} \sim \mathcal{CN}(\bb0, \sigma_0^2 \bbI_{N_r})$ is a vector of complex circular Gaussian noise,  $\bar{\bby} \in \mathbb{C}^{N_r}$ is the received vector, $\bar{\bbx} \in \mathcal{X}^{N_u}$ is the vector of transmitted symbols, which \textit{is the unknown variable},  and $\mathcal{X}$ is a finite set of constellation points.
Although the variables involved are complex, we can rewrite the model in its equivalent real-valued representation obtained by considering the real $\mathfrak{R}(.)$ and imaginary $\mathfrak{I}(.)$ parts separately.
Define $\bbx = [\mathfrak{R}(\bbx)^\top, \mathfrak{I}(\bar{\bbx})^\top]^\top$, $\bby = [\mathfrak{R}(\bar{\bby})^\top ,\mathfrak{I}(\bar{\bby})^\top]^\top$, $\bbz = [\mathfrak{R}(\bar{\bbz})^\top ,\mathfrak{I}(\bar{\bbz})^\top]^\top$ and 
\begin{equation}
\bbH = 
\begin{bmatrix}
    \mathfrak{R}(\bar{\bbH}) & -\mathfrak{I}(\bar{\bbH})\\
    \mathfrak{I}(\bar{\bbH}) & \mathfrak{R}(\bar{\bbH})\\
\end{bmatrix}.
\end{equation}
Thus, the system can be rewritten in the equivalent real-valued representation as in~\eqref{E:mimo_model}.
As we assume that the symbols' prior distribution is uniform among the constellation elements and the measurement noise $\bbz$ is Gaussian, the maximum a posteriori (MAP) detector boils down to a maximum likelihood (ML) detector
\begin{equation}\label{eq:ml}
	\hat{\bbx}_{\mathrm{ML}} = \argmin_{\bbx \in \mathcal{X}^{N_u}}\,\, ||\bby - \bbH\bbx||^2_2,
\end{equation}
The exact solution to this problem is NP-hard due to the discrete nature of the constellation $\mathcal{X}$.
Therefore, approximate solutions have been proposed.
We consider the following existing solutions as baseline methods.

\begin{itemize}
    \item {\bf MMSE}~\cite{Proakis2007}: Linear detector, which formally solves first a relaxation of~\eqref{P:main}, by considering  $\bbx \in \mathbb{C}^{N_u}$ instead of $\bbx \in \ccalX^{N_u}$, and the projects the solution back onto one of the constellation elements.
    \item {\bf V-BLAST}~\cite{Chin2002ParallelMD}: Multi-stage interference cancellation BLAST algorithm using zero-forcing as the detection stage.
    \item {\bf SDR}~\cite{sphered}: Sphere-decoding, a search algorithm that prunes the search space where $||\bby - \bbH \bbx||^2 > r$.
    \item {\bf RE-MIMO}~\cite{remimo}: Recurrent permutation equivariant neural detector based on an encoder-predictor architecture. The encoder and the predictor are parameterized by a transformer and a multi-layer perceptron, respectively.
    We follow the training procedure proposed in the paper.
    \item {\bf OAMPNet}~\cite{oampnet}: Unfolding of the ortoghonal approximate message passing (OAMP) algorithm. 
    We use 10 layers as it is proposed in the paper. 
    At each layer, a matrix pseudo inverse is required and has 2 learnable parameters.
    \item {\bf Overdamped Langevin-based detector~\cite{zilberstein2022}}: 
    Detector based on annealaed version of overdamped Langevin dynamics, where the symbol's estimation is obtained by sampling from the (approximate) posterior distribution.
    We specify the hyperparameters in Table~\ref{table:hyperparam}.
    \item {\bf ML}: The optimal solver for~\eqref{eq:ml} using Gurobi~\cite{gurobi}, a highly-optimized mixed integer programming package.
\end{itemize}

\vspace{1mm}

\noindent The signal-to-noise ratio (SNR) is given by 
\begin{equation}\label{eq:snr}
	\text{SNR} = \frac{\mathbb{E}[||\bbH\bbx||^2]}{\mathbb{E}[||\bbz||^2]}.
\end{equation}

\noindent For all the experiments in the following subsections, we consider a 16-QAM modulation.
The simulation environment includes a base station with $N_r=64$ receiver antennas and $N_u=32$ single-antenna users.
We consider a Kronecker correlated model given by
\begin{equation}
	\bbH = \bbR_r^{1/2}\bbH_e \bbR_u^{1/2},
	\label{E:kron}
\end{equation}
where $\bbH_e$ is a Rayleigh fading channel matrix and $\bbR_r$ and $\bbR_u$ are the spatial correlation matrices at the receiver and transmitters, respectively.
These correlation matrices are generated according to the exponential correlation matrix model with a coefficient $\rho = 0.6$; see \cite{Loyka2001} for details. 
Before going to the experiments, we give the closed-form expression of the constituent terms in~\eqref{E:score_function_spectral}, based on the SVD of the channel matrix given by $\bbH = \bbU\boldsymbol{\Sigma}\bbV^\top$; for details see~\cite{zilberstein2022}.

\vspace{0.4cm}
\noindent {\bf Algorithm and computational complexity.}
We define the spectral representation of $\tilde{\bbx}_l$ and $\bby$ as $\tilde{\boldsymbol{\chi}}_l = \bbV^{\top} \tilde{\bbx}_l$ and $\boldsymbol{\eta} = \bbU^{\top}\bby$, and $s_j$ the $j$-th singular value.
Then, the score of the likelihood is:
\begin{equation}\label{eq:score_likeli}
    \nabla_{\tilde{\boldsymbol{\chi}}_l} \! \log p(\boldsymbol{\eta}|\tilde{\boldsymbol{\chi}}_l, \bbH)  = \boldsymbol{\Sigma}^\top \,\, |\sigma_0^2\bbI - \sigma_l^2\boldsymbol{\Sigma}\boldsymbol{\Sigma}^\top|^{\dagger}\,\, ( \boldsymbol{\eta} - \boldsymbol{\Sigma} \tilde{\boldsymbol{\chi}}_l).
\end{equation}
For the score of the prior [cf.~\eqref{eq:prior}], we need the conditional expectation, which can be calculated elementwise as
\begin{align}\label{E:mixed_gaussian}
	\mathbb{E}_{\sigma_l}[x_j|[\tilde{\bbx}_l]_j] &= \frac{1}{Z}\sum_{x_k \in \ccalX} x_k \exp\bigg(\frac{-||[\tilde{\bbx}_l]_j - x_k||^2}{2\sigma_l^2}\bigg),
\end{align}
where $Z = \sum_{x_k \in \ccalX} \exp\Big(\frac{-||[\tilde{\bbx}_l]_j - x_k||^2}{2\sigma_l^2}\Big)$ and $j=1,\cdots, N_u$. 
Then, given the orthogonality of $\bbV$, we have $\nabla_{\tilde{\boldsymbol{\chi}}_l}\log p(\tilde{\boldsymbol{\chi}}_l) = \bbV^{\top}\nabla_{\tilde{\bbx}_l}\log p(\tilde{\bbx}_l)$.

Additionally, given that the sample after the annealing process will be very close to the constellation but not exactly, we take $\bar{\bbx} = \argmin_{\bbx \in \ccalX^{N_u}}||\bbx - \bbV\tilde{\boldsymbol{\chi}}_{T,L}||_2^2$. 
In our implementation, we generate $U$ different Langevin samples $\{\bar{\bbx}_u\}_{u=1}^U$ for each pair $\{\bby, \bbH\}$ by running Algorithms~\ref{alg_uld} and~\ref{alg_3ld} multiple times and keep the sample that minimizes
\begin{equation}\label{eq:Ntraj}
    \hat{\bbx} = \argmin_{\bbx \in \{\bar{\bbx}_u\}_{u=1}^U} ||\bby - \bbH\bbx||_2^2.
\end{equation}
\noindent 
Notice that these $U$ Langevin trajectories can be run in parallel, as they are independent of each other. We fix $U = 20$ as it was explained in~\cite{zilberstein2022}.

We set $\gamma = 1$, and for the third-order, $\lambda = 1$ and $\alpha = 1.2$.
Moreover, we consider the same $\epsilon$ for all the noise levels, given by $\epsilon = \frac{\epsilon_0}{\sigma_L^2}$.
Lastly, the pre-condition matrix is a diagonal matrix that depends on the noise level $l$, and is given by
\begin{align}
     [\bbC_l]_{jj} =
		\begin{cases}
		  \sigma_l^2 \left(1 - \frac{\sigma_l^2}{\sigma_0^2}s_j^2\right) \hspace{8mm} \text{if} \,\,\, \sigma_ls_j \leq \sigma_0 \\
		\sigma_l^2 - \frac{\sigma_0^2}{s_j^2} \hspace{10mm}  \text{if} \,\,\, \sigma_ls_j > \sigma_0
		\end{cases}
\end{align}	
and the mass as $\bbM_l = \frac{\gamma^2}{4} \bbC_l^{-1}$.
This choice of $\bbM$ is a natural way of relating the pre-conditioning matrix with the covariance matrix of the marginal distribution of the momentum variable $\bbv_t$ in the invariant distribution defined in~\eqref{eq:inv_distr}; recall that the covariance matrix gives information about the stretching of each direction in the equilibrium case.
Notice that $\bbM$ also appeared in the marginal distribution of the auxiliary variables in the higher-order dynamics defined in Section~\ref{subsec:pre_cond}.
Now, we present the results of the experiments.
In the three experiments in this section, we consider the ABO numerical method for the underdamped and the (BC)OA(BC) for the third-order algorithm.

Regarding the computational complexity, for MIMO detection we first compute the SVD of the channel $\bbH$, whose complexity is 
\linebreak$\ccalO(N_uN_r\min\{N_u, N_r\})$, and is done only once per channel.
Then, the discretization schemes entail three steps.
The first and the third steps are just vector summations, since we consider the mass parameter $\bbXi = \xi \bbI$ to be a scalar in this work. 
The second step requires the computation of~\eqref{eq:score_likeli} and~\eqref{E:mixed_gaussian}, which entail a complexity of $\ccalO(N_u^2 + KN_u)$ per iteration.
Therefore, the overall complexity, including the SVD computation and all the iterations, is $\ccalO(N_uN_r\min\{N_u, N_r\} + LT(N_u^2 + KN_u))$.
Compared to the overdamped case (see~\cite{zilberstein2022}), we see that we are not adding order complexity to the detector.
Regarding the $U$ trajectories (Langevin samples), observe that these are independent of each other, so they can be computed in parallel.
Hence, we consider a single trajectory in the above complexity analysis.

\begin{figure*}[t]
    \centering
	\begin{subfigure}{.3\textwidth}
    	\centering
    	\includegraphics[width=1\textwidth]{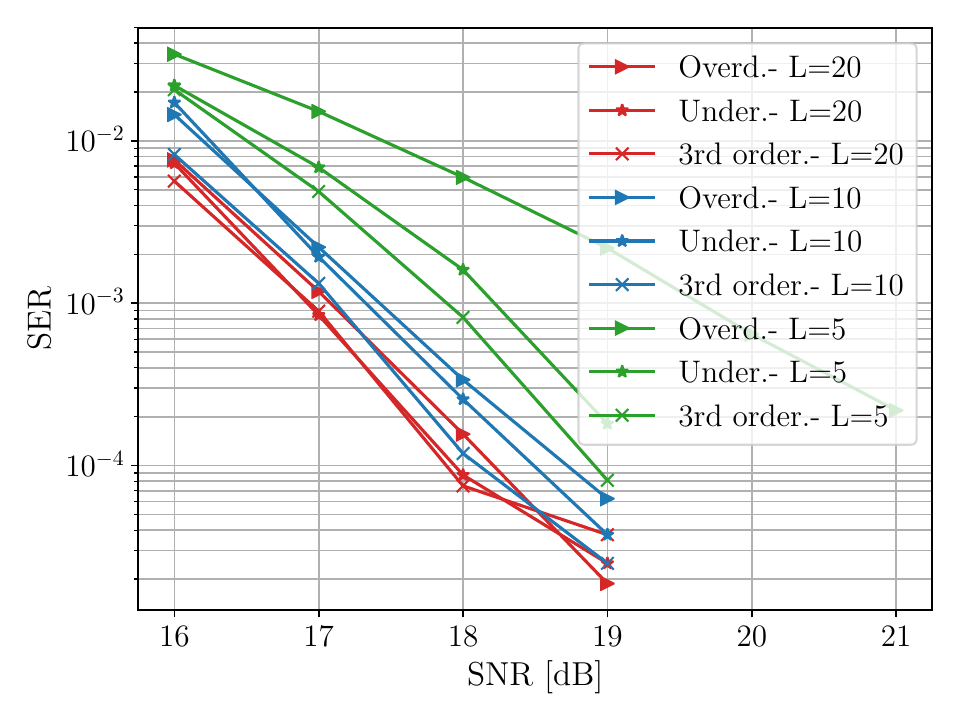}
    	\vspace{-0.15in}
    	\caption{}
    	\label{fig:SER-noiselevels}
	\end{subfigure}%
	\centering
	\begin{subfigure}{.3\textwidth}
    	\centering
    	\includegraphics[width=1\textwidth]{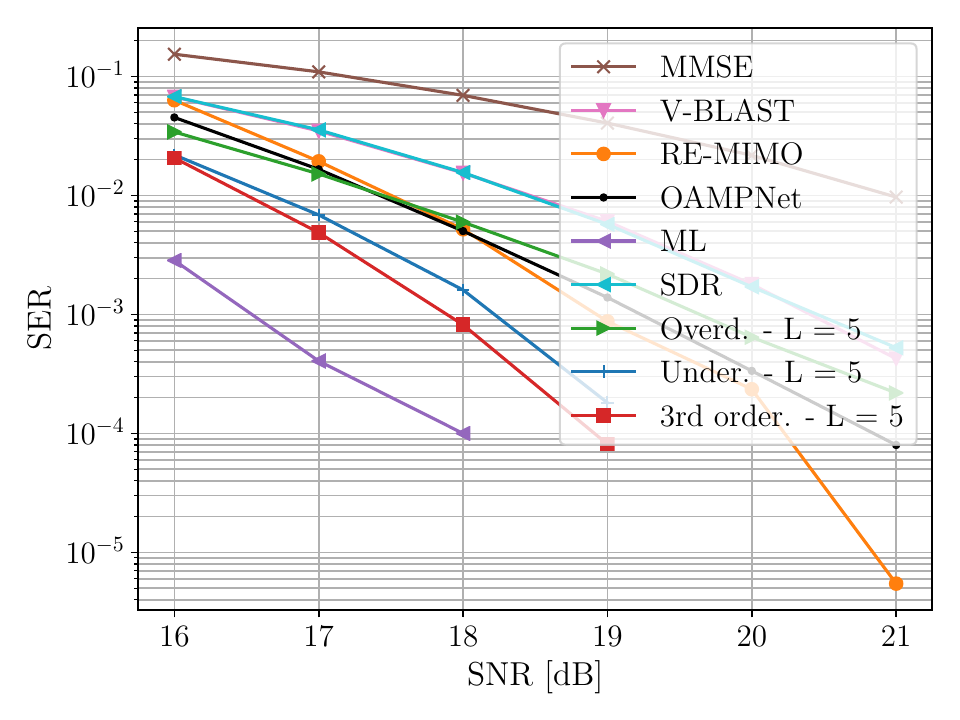}
    	\vspace{-0.15in}
    	\caption{}
    	\label{fig:SER-methods-detection}
	\end{subfigure}
	\centering
	\begin{subfigure}{.3\textwidth}
    	\centering
    	\includegraphics[width=1\textwidth]{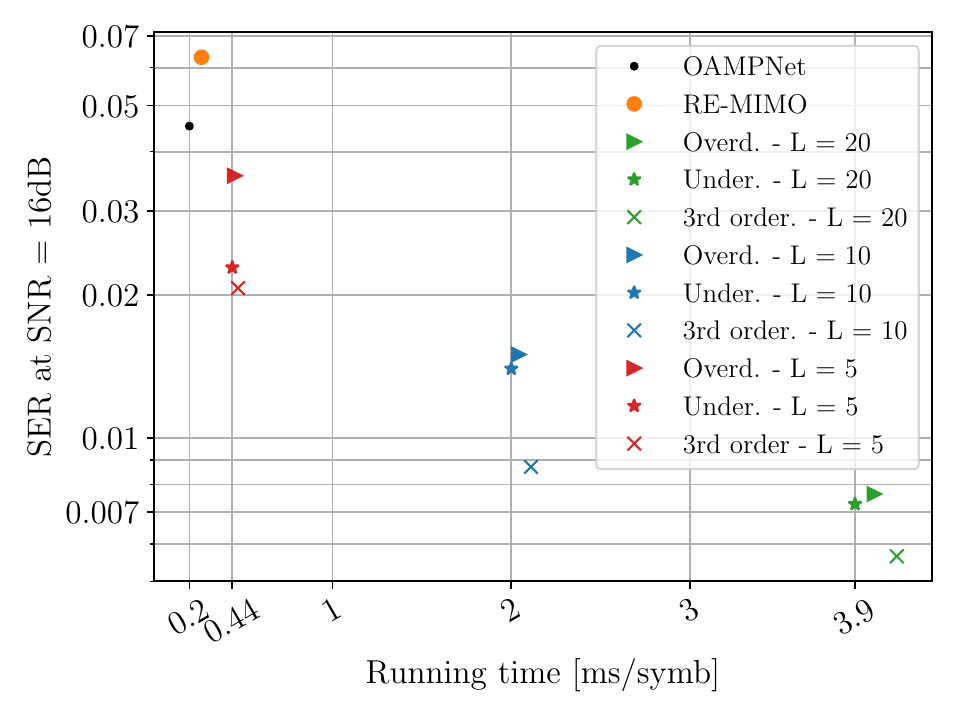}
    	\vspace{-0.15in}
    	\caption{}
    	\label{fig:SER-complexity-detection}
	\end{subfigure}
	\vspace{-0.02in}
	\caption{ {\small Performance analysis of our proposed methods for MIMO symbol detection considering SER as a function of SNR. 
	(a)~Comparison between our annealed underdamped and third-order Langevin method and the overdamped variant for $L \in \{5, 10, 20\}$. 
	(b)~Comparison with classical and learning-based detection methods. 
	(c)~Comparison with different detection methods for an $\text{SNR}=\SI{16}{\decibel}$ and varying the running time in $\mathrm{ms}/\mathrm{symb}$. We consider only one trajectory $U=1$ for both annealed underdamped and third-order Langevin.}}
	\vspace{-0.1in}
	\label{figs_analysis_detector}
\end{figure*}

\vspace{0.4cm}
\noindent {\bf Comparison with the overdamped Langevin detector.} 
In this experiment, we compare the advantage of using the pre-conditioned annealed underdamped and third-order Langevin detector w.r.t. the overdamped case. 
We consider three cases where $L \in \{5, 10, 20\}$.
All hyperparameters are summarized in Table~\ref{table:hyperparam} and the results are shown in Fig.~\ref{fig:SER-noiselevels}.
First, we observe that when $L=20$, all methods have comparable performance. 
On the other hand, for both $L=10$ and $L=5$, our proposed methods outperform the overdamped Langevin detector, with the third-order based detector outperforming the underdamped case.
Notice that the number of iterations ($L \times T$) was reduced by a factor of 10 (150 vs 1400) and 2 for $L=5$ and $L=10$, respectively.
Therefore, this experiment illustrates that there is a \emph{trade-off between performance and running time} and that \emph{adding auxiliary variables entails a better performance with a reduced running time.}

\begin{table*}[t]
    \centering
    \caption{Hyperparameters of the different algorithms based on Langevin dynamics.}
    \begin{tabular}{c|c c c|c c c | c c c}
        Parameter & \multicolumn{3}{c }{Overdamped} & \multicolumn{3}{c }{Underdamped} & \multicolumn{3}{c }{Third-order} \\ 
        \hline
        $L$ &  $5$ & $10$ & $20$ &
                $5$ & $10$ & $20$ & 
                $5$ & $10$ & $20$ \\ 
        
        $\sigma_1$ &
                $0.4$ & $1$ & $1$ & 
                $0.4$ & $1$ & $1$ & 
                $0.4$ & $1$ & $1$ \\ 
        
        $\sigma_L$ &
                $0.02$ & $0.01$ & $0.01$  & 
                $0.02$ & $0.01$ & $0.01$  & 
                $0.02$ & $0.01$ & $0.01$ \\ 
        $\epsilon_0$ &
                $6\times 10^{-4}$ & $3\times 10^{-5}$ & $3\times 10^{-5}$ & 
                $6\times 10^{-4}$ & $3\times 10^{-5}$ & $3\times 10^{-5}$ & 
                $2.2\times 10^{-4}$ & $5\times 10^{-5}$ & $5\times 10^{-5}$ \\ 
        $T$ &
                $30$ & $70$ & $70$ & 
                $30$ & $70$ & $70$ &
                $30$ & $70$ & $70$ \\ 
        $\tau$ &
                $0.01$ & $0.5$ & $0.5$ & 
                $0.01$ & $0.5$ & $0.5$ & 
                $0.023$ & $0.084$ & $0.084$ \\ 
        
    \end{tabular}
    \label{table:hyperparam}
\end{table*}

\vspace{0.4cm}
\noindent {\bf Comparison with other methods.} 
Based on our previous experiments, we consider the case of $L=5$ for all the methods as it is the case with the best trade-off between running time and performance.
The comparison with baseline detectors is shown in Fig.~\ref{fig:SER-methods-detection}.
The figure reveals that our proposed method markedly outperforms the other detectors.
Notice that our method can handle a varying number of users without the need for any retraining as required in, e.g., OAMPNet~\cite{oampnet}.
This is key in MIMO communications, as the number of users connected to the network might be constantly changing.

\vspace{0.4cm}
\noindent {\bf Running time comparison.} 
In this third experiment, we compute the SER of our method and the other baselines w.r.t. running time in $\mathrm{ms}/\mathrm{symb}$.
We assume a coherence time such that each block contains 1000 samples.
The comparison is shown in Fig.~\ref{fig:SER-complexity-detection} for an $\text{SNR}= \SI{16}{\decibel}$, $5000$ symbols, and only one trajectory, i.e., $U=1$ in~\eqref{eq:Ntraj}; this means that we are considering a full pass of both Alg.~\ref{alg_uld} and Alg.~\ref{alg_3ld}.
We focus on $U=1$ because all trajectories can be computed in parallel, thus, larger values for $U$ result in similar wall-clock computational time.
Given the coherence time, we have to compute 5 SVDs that correspond to each $\{\bbH_i\}_{i=1}^5$.
First, notice that both underdamped and overdamped methods have the same performance when considering $L = 20$, something expected given the result in the first experiment.
Moreover, the third-order slightly outperforms both.
However, when $L =5$, both the underdamped and the third-order cases successfully reduce the running time while achieving a better performance compared to the overdamped case.
Furthermore, the third-order outperforms the underdamped without adding computational burden.

\subsection{Channel estimation with learned score prior}
\label{subsec:channel_estimation}

We analyze the proposed methods for the problem of MIMO channel estimation.
Again, we cast this problem as in~\eqref{E:mimo_model}, where $N_u$ are the number of single-antenna transmitters or users and $N_r$ the antennas at the base station.
The channel matrix $\bbH \in \mathbb{C}^{N_r \times N_u}$ is \textit{the unknown variable}, $\bbZ$ is a random matrix where each column $\bbz \sim \mathcal{CN}(\bb0, \sigma_0^2 \bbI_{N_r})$ is a vector of complex circular Gaussian noise, and $\bbY \in \mathbb{C}^{N_r \times N_p}$ is the received vector.
In this problem, the variable $\bbx$ is a matrix denoted as $\bbP \in \ccalX^{N_u \times p}$, which are known as pilots, and typically they are pre-specified.
Thus, the forward model is $\bbY = \bbH\bbP + \bbZ$, and the optimal solution in~\ref{eq:map}
is 
\begin{align}\label{eq:map_H}
	\hat{\bbH}_{\mathrm{MAP}}
	= \argmax_{\bbH \in \mathbb{C}^{N_r\times N_u}}\,\, p_{\bbz}(\bbY - \bbH\bbP)p(\bbH).
\end{align}
Similar to the case of symbol detection, the problem of channel estimation can also be rewritten to treat the real and imaginary parts separately~\cite{arvinte2022}, boiling down to the expression in~\eqref{E:mimo_model}.
The dimension $N_p = \alpha_p N_u$ denotes the number of pilots, and the value of $\alpha_p$ determines if the problem is ill-conditioned or not; if $\alpha < 1$, then the problem is underdetermined.
We consider $\alpha_p = 0.6$.
Thus, it becomes necessary to incorporate prior knowledge about the channel distribution to obtain an accurate estimation. 
We describe the methods that we use as baselines.

\begin{itemize}
    \item {\bf Lasso}~\cite{lasso}: This is a compressive-sensing-based approach that uses $\ell_1$-norm element-wise regularization in the two-dimensional Fourier (beamspace) domain.
    \item {\bf fsAD}~\cite{fsad}: This is another classical compressive-sensing-based approach for recovering channel matrices assumed sparse in the continuous spatial frequency domain, similar to Newtonized OMP (orthogonal matching pursuit)~\cite{fsad2}.
    \item {\bf L-DAMP}~\cite{ldamp1, ldamp2}: This method is a data-driven algorithm that uses algorithm unrolling~\cite{eldar2021} and end-to-end learning. A different model is trained for each SNR.
    \item {\bf Overdamped Langevin-based detector}~\cite{arvinte2022}: This is the annealed first-order Langevin dynamic.
    The score of the prior is learned from data using denoising score-matching.
\end{itemize}

\noindent In all the experiments, we consider the SNR as in~\eqref{eq:snr} and the estimation quality as the normalized mean squared error (NMSE)
\begin{equation}
    \text{NMSE[dB]} = 10\log_{10}\frac{||\hat{\bbH} - \bbH||_F^2}{||\bbH||_F^2}.
\end{equation}
For the channel model, we use the standardized clustered delay line (CDL)-C model, which is adopted in 5G specifications~\cite{cdlc}.
We consider $N_u = 64$ and $N_r = 16$.

\vspace{0.4cm}
\noindent {\bf Algorithm and computational complexity.}
In this case, the score of the likelihood is:
\begin{equation}\label{eq:score_likeli_H}
    \nabla_{\tilde{\bbH}_l} \! \log p(\bbY|\tilde{\bbH}_l, \bbP)  \approx \frac{(\bbY - \tilde{\bbH}_l\bbP)\bbP^{\text{H}}}{\sigma_0^2 + \sigma_l^2}.
\end{equation}
For the score of the prior [cf.~\eqref{eq:prior}], we parameterize it with a neural network, known as score \emph{network}, and train using denoising score-matching~\cite{ermon2019}.
We use the architecture proposed in~\cite{ermon2020}. 
We trained the model as it is explained in~\cite{arvinte2022}, and we use the same score network for our higher-order methods.

For the step-size, we consider $\epsilon_l = \epsilon_0\frac{\sigma_l^2}{\sigma_L^2}$.
For the underdamped case we consider $\gamma = 1$, while for the third-order case we set $\lambda = 1$ and $\alpha = 2$.
Moreover, we consider $\epsilon_0= 2.2\times 10^{-10}$, $L = 58$, and $T = 3$ for both methods, and an early stopping at the iteration 60.
The mass matrix depends on the SNR level: we consider $\bbM = 3\bbI$ for the first two levels of SNR and $\bbM = 2\bbI$ for the remaining levels.
Lastly, the pre-condition matrix is not used, i.e., $\bbC_l = \bbI$.

The computational complexity depends on the neural network in this case, which is explained in~\cite{arvinte2022}.
However, similar to the experiment in Section~\ref{subsec:detection}, both algorithms~\ref{alg_uld} and~\ref{alg_3ld} do not add computational burden.

\begin{figure*}[t]
    \centering
	\begin{subfigure}{.3\textwidth}
    	\centering
    	\includegraphics[width=1\textwidth]{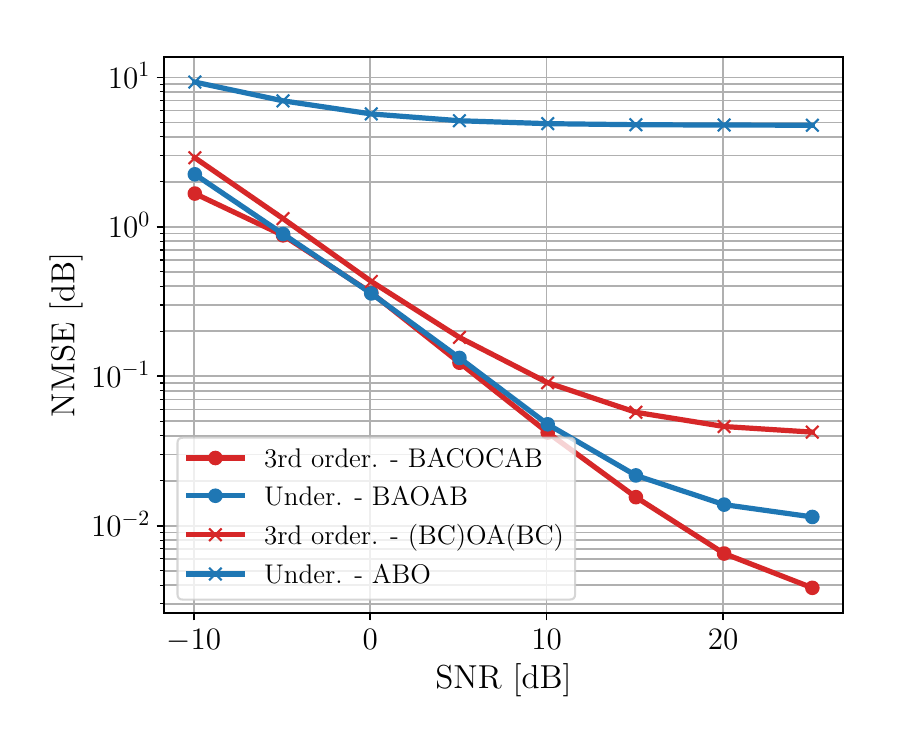}
    	\vspace{-0.15in}
    	\caption{}
    	\label{fig:NMSE-discretization_methods}
	\end{subfigure}%
	\centering
	\begin{subfigure}{.3\textwidth}
    	\centering
    	\includegraphics[width=1\textwidth]{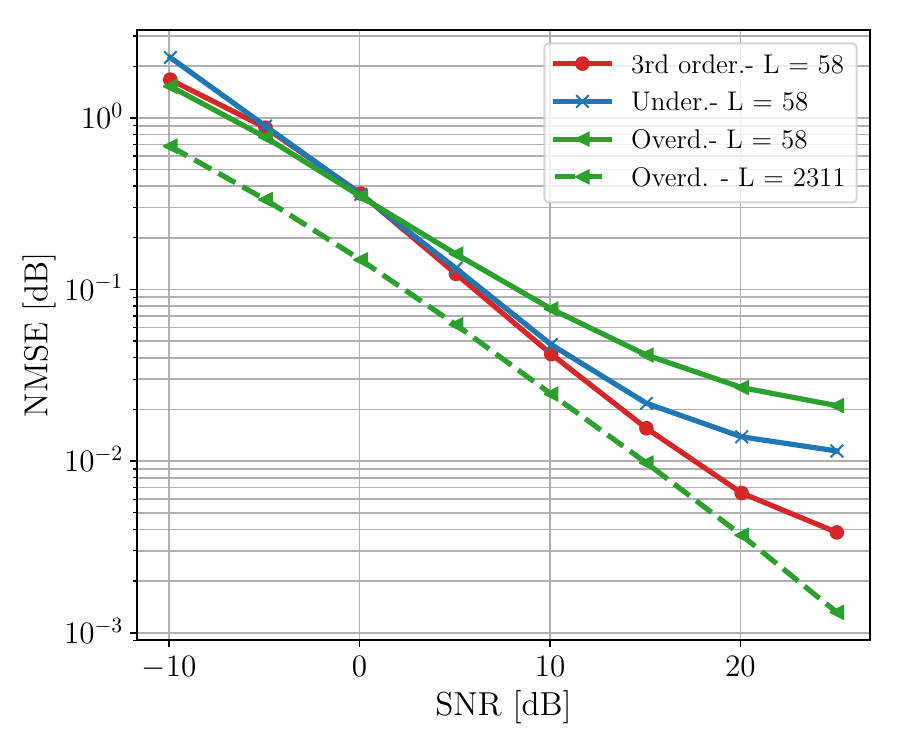}
    	\vspace{-0.15in}
    	\caption{}
    	\label{fig:NMSE-langevin-order}
	\end{subfigure}
	\centering
	\begin{subfigure}{.3\textwidth}
    	\centering
    	\includegraphics[width=1\textwidth]{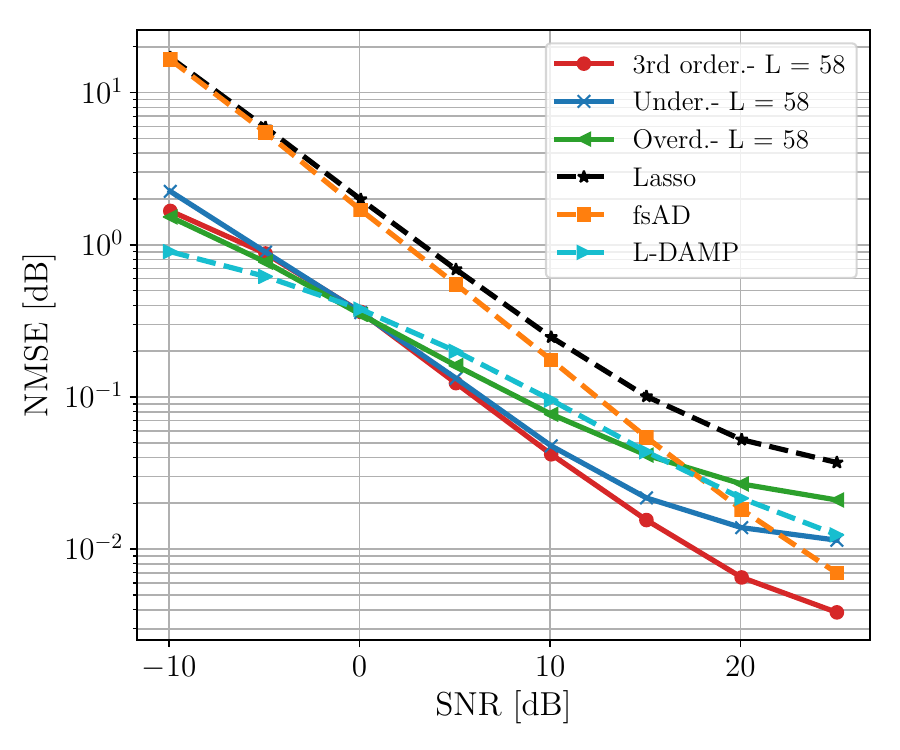}
    	\vspace{-0.15in}
    	\caption{}
    	\label{fig:Hestimation_methods}
	\end{subfigure}
	\vspace{-0.02in}
	\caption{ {\small Performance analysis of our proposed methods for channel estimation considering NMSE as a function of SNR in a CLD-C channel model. 
	(a)~Comparison of our proposed annealed underdamped and third-order Langevin methods using different discretization schemes. 
	(b)~Comparison of our proposed annealed underdamped and third-order Langevin method and the annealed overdamped Langevin.
	(c)~Comparison with baseline methods.}}
	\vspace{-0.1in}
	\label{figs_channel_estimation}
\end{figure*}

\vspace{0.4cm}
\noindent {\bf Comparison between discretization schemes.} 
In this first experiment, we compare the performance of the annealed underdamped and third-order algorithms when considering different discretization schemes.
In particular, we consider the methods introduced in Section~\ref{subsec:CLDdiscr}, i.e., ABO and BAOAB for ULD, and (BC)OA(BC) and BACOCAB for the third-order dynamic.
The results are shown in Fig.~\ref{fig:NMSE-discretization_methods}.
First, notice that the annealed underdamped algorithm is unstable when considering the ABO discretization, while it becomes stable when considering the BAOAB method.
In the case of the third-order dynamic, the results reveal that it is more robust to the choice of numerical discretization.
However, we observe that the BACOCAB outperforms the (BC)OA(BC), in particular for SNRs higher than \SI{0}{\decibel}.
Overall, we see that in both cases, propagating more times each sub-operator within one discretization interval entails a better performance.

\vspace{0.4cm}
\noindent {\bf Comparison with the overdamped Langevin detector.} 
In this second experiment, given the discretizations BAOAB and BACOCAB for the underdamped and third-order respectively, we compare their performance with the overdamped case proposed in~\cite{arvinte2022}, which uses $L=2311$.
Also, we add the case of $L = 58$, with the same hyperparameters as the underdamped and third-order case.
The results are shown in Fig.~\ref{fig:NMSE-langevin-order}.
Although the overdamped case with $L=2311$ outperforms our proposed methods, the difference compared to the third-order case is small for SNRs higher than $\SI{0}{\decibel}$. 
For lower SNRs, all the methods have a similar performance.
There is a trade-off between running time and performance, and which to prioritize depends on the accuracy of the estimation that is needed.

\vspace{0.4cm}
\noindent {\bf Comparison with other methods.} 
Based on our previous experiments, we consider the case of $L=58$ for all the methods as it is the case of less complexity within all the combinations that we proposed and compare with the baseline detectors.
The comparison is shown in Fig.~\ref{fig:Hestimation_methods}.
We observe that our proposed methods outperform the other baselines for SNRs greater than $\SI{0}{\decibel}$.
Thus, despite the trade-off that was shown in the previous experiment where our proposed annealed third-order algorithm performs slightly worse than the overdamped case, it still outperforms the other methods.

\subsection{Inverse problem in imaging with learned score prior}
\label{subsec:image_estimation}

Finally, we analyze the proposed methods for three inverse problems in imaging, namely image deblurring, inpainting, and super-resolution.
In this context, the degradation operator $\bbH$ is known, and we want to estimate the signal given a noisy observation.
Thus, the problem is equivalent to~\eqref{eq:map}.
Similar to the experiment in Section~\ref{subsec:channel_estimation}, the problem is ill-conditioned, and therefore, we need to regularize the problem by incorporating prior information.
We do this by using the score network proposed in~\cite{ermon2020} as well as the pre-trained model on the FFHQ dataset~\cite{karras2019style}.
We consider 100 validation samples at random, and the following methods as baseline
\begin{itemize}
    \item {\bf DDRM}~\cite{kawar2022denoising}: Denoising diffusion restoration model, which is based on the denoising diffusion probabilistic model~\cite{ho2020denoising}. We use the pre-trained model from~\cite{choi2021ilvr}. We consider the same setting proposed by the authors.
    \item {\bf SNIPS}~\cite{kawar2021snips}: This is the annealed first-order Langevin dynamic. For the score network, we use the pre-trained model from~\cite{ermon2020}. We consider 20 levels of noise, with 3 levels per sample, $\sigma_1 = 348$ and $\sigma_L = 0.01$.
    \item {\bf DPS}~\cite{chung2023diffusion}: Solver of general noisy (non)linear inverse problems based on diffusion and an approximation of the posterior sampling. To be fair in the comparison, we consider 60 noise levels, which corresponds to the same number of evaluations of the NN as Langevin with 20 levels of noise and 3 samples per level. We use the pre-trained model given by the authors.  
\end{itemize}

We generate only one sample with SNIPS and our two proposed methods, similar to diffusion-based models.
As evaluation metrics, we consider PSNR, LPIPS~\cite{zhang2018unreasonable}, and FID~\cite{heusel2017gans}.
See the supplementary material to see the images obtained for each task and the hyper-parameters of our methods.
The results for noiseless experiments -- $\sigma_0 = 0.0002$ -- are in Table~\ref{table:results_noiseless_images}, while for noisy case -- $\sigma_0 = 0.05$ -- in Table~\ref{table:results_noisy_images}.

\vspace{0.4cm}
\noindent {\bf Image deblurring with anisotropic Gaussian filter.} 
In this first experiment, we consider the anisotropic Gaussian blur kernel with $\sigma = 20$ horizontally and $\sigma = 1$ vertically, which was used in~\cite{kawar2022denoising}. 
We observe in Table~\ref{table:results_noiseless_images} that for the noiseless case, the algorithm based on 3rd-order Langevin outperforms the other baselines.
Furthermore, we observe a similar behavior than the experiments in Section~\ref{subsec:channel_estimation}: when the SNR between the signal and the noise is high -- noiseless case --, then the performance when considering higher-order methods increases as well.
However, when the SNR is low -- noisy case -- the three dynamics achieve similar performance. 
In this case, the 3rd order Langevin sampler outperforms all the other methods in LPIPS and FID, while for the PSNR DDRM achieves the best performance.
Lastly, we show in Figs. 1 and 2 of the supplementary material some images generated with our proposed methods.
We observe that for the noiseless case, the image generated with the 3rd order Langevin sampler is the less blurry image, while for the noisy case, the three images from the different Langevin samplers look similar.

\vspace{0.4cm}
\noindent {\bf Image inpainting.} 
For inpainting, we randomly drop 50 of the pixels from the original image.
Again, the algorithm based on 3rd order Langevin outperforms the other baselines as we show in Tables~\ref{table:results_noiseless_images} and~\ref{table:results_noisy_images}
We can see also that for the noisy case the three Langevin dynamics perform similar, while for the noiseless case, the gap in performance increases.
Notice also that our proposed method outperforms DDRM in the noisy setting. 
Lastly, we show in Figs. 3 and 4 of the supplementary material some images generated with our proposed methods.

\vspace{0.4cm}
\noindent {\bf Super-resolution.} 
Lastly, we consider super-resolution, using a block averaging filter of 4 on each axis.
In this case, DDRM achieves the best performance in terms of PSNR, although the 3rd-order Langevin sampler achieves a slightly better LPIPS in the noiseless case.

\begin{table*}[t]
    \centering
    \caption{{\small Quantitative results on noiseless images ($\sigma_0 = 0.0002$) from FFHQ. We consider the average PSNR[dB], LPIPS and FID of Gaussian deblurring, inpainting and $4\times\text{SR}$. In bold is the best method for each metric and experiment.}}
    \begin{tabular}{c | c c c|c c c | c c c}
         & \multicolumn{3}{c }{Deblurr (Gauss)} & \multicolumn{3}{c }{Inpainting} & \multicolumn{3}{c }{SRx4} \\ 
        \hline
        Method & PSNR [dB] $\uparrow$ & LPIPS $\downarrow$ & FID $\downarrow$ &  PSNR [dB] $\uparrow$ & LPIPS $\downarrow$ & FID $\downarrow$ &  PSNR [dB] $\uparrow$ & LPIPS $\downarrow$ & FID $\downarrow$ \\ 
        \hline 
        DDRM & 35.82 & 0.017 & 13.81 &  30.77 & 0.04 & 38.74 & \textbf{28.53} & 0.13 & \textbf{58.06} \\ 
        DPS & 18.64 & 0.36 & 110.65 &  19.06 & 0.35 & 117.27   & 15.64 & 0.45 & 120.38 \\ 
        SNIPS & 34.76 & 0.009 & 22.51 &  30.68 & 0.035 & 35.2 & 28.18 & 0.15 & 68.05 \\ 
        Underd. Lang &  35.55 & 0.008 & 16.34 &  30.98 & 0.026 & 29.17 & 28.22 & 0.13 & 61.41 \\ 
        3rd. order Lang & \textbf{36.78} & \textbf{0.004} & \textbf{11.22} & \textbf{31.46} & \textbf{0.02} & \textbf{21.65} & 28.41 & \textbf{0.12} & 58.81 \\

    \end{tabular}
    \label{table:results_noiseless_images}
\end{table*}

\begin{table*}[t]
    \centering
    \caption{{\small Quantitative results on noisy ($\sigma = 0.05$) images from FFHQ. We consider the average PSNR[dB], LPIPS and FID of Gaussian deblurring, inpainting and $4\times\text{SR}$. In bold is the best method for each metric and experiment.}}
    \begin{tabular}{c|c c c|c c c | c c c}
         & \multicolumn{3}{c }{Deblurr (Gauss)} & \multicolumn{3}{c }{Inpainting} & \multicolumn{3}{c }{SRx4} \\ 
        \hline
        Method &  PSNR [dB] $\uparrow$ & LPIPS $\downarrow$ & FID $\downarrow$ & PSNR [dB] $\uparrow$ & LPIPS $\downarrow$ & FID $\downarrow$ &  PSNR [dB] $\uparrow$ & LPIPS $\downarrow$ & FID $\downarrow$\\ 
        \hline 
        DDRM & \textbf{28.25} & 0.175 & 84.09 &  29.2 & 0.09 & 52.8 &  \textbf{27.02} & \textbf{0.187} & \textbf{90.54} \\ 
        DPS & 18.14 & 0.38 & 114.32 &  18.64 & 0.37 & 115.7 & 15.53 & 0.46 & 117.67 \\ 
        SNIPS & 27.2 & 0.21 & 92.5 &  29.54 & 0.077 & 57.61 & 26.33 & 0.22 & 104.17     \\ 
        Underd. Lang & 27.06 & 0.18 & 89.14 &  29.4 & 0.06 & 52.49 & 26.16 & 0.21 & 100.61 \\ 
        3rd. order Lang & 27.29 & \textbf{0.173} & \textbf{82.83} &  \textbf{29.73} & \textbf{0.054} & \textbf{44.13} & 26.36 & 0.191 & 94.17 \\ 
        
    \end{tabular}
    \label{table:results_noisy_images}
\end{table*}

\section{Conclusions and Future Work}
\label{sec:conclusions}
In this paper, we proposed a general framework for solving linear inverse problems based on a pre-conditioned annealed version of higher-order Langevin dynamics, namely the underdamped and third-order diffusion models.
We formally defined the continuous-time dynamic with pre-conditioning and characterized their corresponding invariant distributions.
Based on these continuous-time dynamics, we derived two algorithms using splitting techniques for discretization.

Revisiting the motivating question in Section~\ref{sec:problem_formulation}, we demonstrated that adding auxiliary variables and designing pre-conditioning matrices and/or discretization schemes entail an accelerated solver for linear inverse problems.
We show this in two different problems: MIMO symbol detection and channel estimation.
Furthermore, our framework allows us to incorporate both learning-based prior information (as in our channel estimation example) as well as a closed-form expression for the prior (as in MIMO detection).
Through extensive analysis and experiments, we have demonstrated that the high-order methods yield a solution that strikes a balance between running time and performance and is general enough to handle different problems.

A main limitation of the proposed method is the number of hyperparameters to tune. 
A possible research direction to automatically select the hyperparameters is incorporating the dynamic in an unrolling framework~\cite{eldar2021, UWMMSE, Boningpaper} and learning the hyperparameters from data.
From the theoretical point of view, future work includes deriving non-asymptotic guarantees as well as considering a more general memory kernel in GLE.
Finally, from the experimental side, we plan to apply this framework to the joint estimation~\cite{chung2022parallel} of the linear operator $\bbH$ and the hidden signal $\bbx$. 

\vspace{-1mm}
\begin{appendices}

\section{Proof of Proposition~\ref{prop1}}\label{A1}

We define the following matrices
\begin{equation}
    \bbD = \begin{bmatrix}
        0 & 0 \\
        0 & \gamma \bbM
    \end{bmatrix},
    \hspace{1cm}
    \bbQ_t = \begin{bmatrix}
        0 & -\bbC_t \\
        \bbC_t & 0
    \end{bmatrix}.
\end{equation}
Given these matrices, we rewrite the SDE in~\eqref{eq:ct_underdamped_langevin_precond} as follows
\begin{equation}\label{eq:sde_ma}
    \text{d}\bbX_t = -(\bbD + \bbQ_t)\nabla H(\bbX_t)\text{d}t + \sqrt{2\tau\bbD}\, \text{d}\bbW_t.
\end{equation}
In this way, we have boiled down our dynamic to the general SDE formulation considered in~\cite{ma2015complete}.
Thus we can apply~\cite[Theorem 1]{ma2015complete} to show that the invariant distribution of the process is given by~\eqref{eq:inv_distr}.
Furthermore, leveraging our Assumptions~\ref{Ass1} and~\ref{Ass2}, we have that $(\bbD + \bbQ_t)\nabla H(\bbX_t)$ is \textit{Lipschitz continuous for all} $t$, so we can apply~\cite[Theorem 5.2.1]{oksendal2003stochastic} to show that it is unique.
Hence, the only thing left to show is that the Hamiltonian of the pre-conditioned ULD is given by $H(\bbx, \bbv) = U(\bbx) + \frac{1}{2} \bbv^\top \bbM^{-1}\bbv$. We show this next.

We start by writing the Fokker-Planck equation of~\eqref{eq:ct_underdamped_langevin_precond},
which describes the transition probability density of $[\bbx_t, \bbv_t]$ denoted by $\rho(\bbx,\bbv, t)$, and is given by
\begin{equation}\label{eq:fokker_plank}
    \frac{\partial \rho}{\partial t} = \nabla \cdot \left( -\bbb(\bbx,\bbv, t)\rho + \frac{1}{2} \nabla \cdot \left(\boldsymbol{\Sigma}(\bbx,\bbv)\rho\right)\right).
\end{equation}
The term $\bbb(\bbx,\bbv, t)$ is the drift vector field, $\boldsymbol{\Sigma}(\bbx,\bbv,t)$ is the diffusion matrix, which is symmetric and non-negative.
Then, the drift vector field is given by the deterministic components of the SDE in~\eqref{eq:sde_ma}
\begin{equation}
    \bbb(\bbx,\bbv, t) = 
    \begin{bmatrix}
        \bbC_t\bbM^{-1}\bbv\\
        -\bbC_t\nabla U(\bbx) - \gamma\bbv
    \end{bmatrix}
\end{equation}
and the diffusion matrix is given by the matrix that multiplies the random part in~\eqref{eq:sde_ma}
\begin{equation}
    \boldsymbol{\Sigma}(\bbx,\bbv, t)
    = 
    \begin{bmatrix}
        \bb0 \\
        2\gamma \tau \bbM
    \end{bmatrix}
\end{equation}
Thus, we have
\begin{align}
    \frac{\partial \rho}{\partial t} &= \left[\nabla_\bbx, \nabla_\bbv\right] \cdot \bigg(-    
    \begin{bmatrix}
         \bbC_t\bbM^{-1}\bbv\\
        -\bbC_t\nabla U(\bbx) - \gamma\bbv
    \end{bmatrix}\rho + \nonumber \\ \nonumber 
    &\hspace{5cm}\frac{1}{2} \nabla \cdot \left( \begin{bmatrix}
        \bb0 \\
        2\gamma \tau \bbM
    \end{bmatrix}\rho\right)\bigg) \\
    &= \underbrace{\nabla_\bbv \cdot \left(\bbC_t\nabla U(\bbx)\rho\right) - \nabla_\bbx\cdot \left(\bbC_t\bbM^{-1}\bbv \rho\right)}_{\text{Liouville operator of Hamiltonian dynamics}} + \label{eq:FP_ULD} \\\nonumber 
    &\hspace{3cm}\underbrace{\nabla_\bbv \cdot \left(\gamma\bbv\rho\right) + \frac{1}{2} \nabla_\bbv \cdot \nabla_\bbv
        \left(2\gamma \tau \bbM \rho\right)}_{\ccalF_0 \text{of the Ornstein–Uhlenbeck process}}
\end{align}
To show that $\pi(\bbx, \bbv)$ defined in~\eqref{eq:inv_distr} is the invariant distribution, we need to show that it belongs to the kernel of the Fokker-Planck operator.
Consequently, we replace $\rho(\bbx,\bbv,t) = \pi(\bbx, \bbv)$ in~\eqref{eq:FP_ULD}; we start with the first term of~\eqref{eq:FP_ULD}
\begin{align}
    \nabla_\bbv \cdot &\left(\bbC_t\nabla U(\bbx)\rho\right) - \nabla_\bbx\cdot \left(\bbC_t\bbM^{-1}\bbv \rho\right) \\\nonumber
    &= \left(\bbC_t\nabla U(\bbx)\right)^\top \cdot \nabla_\bbv \left(\rho\right) - \left(\bbC_t\bbM^{-1}\bbv\right)^\top \cdot \nabla_\bbx\left(\rho\right) & \\\nonumber
    &= -(\bbC_t\nabla U(\bbx))^\top \cdot \left(\tau^{-1}\bbM^{-1}\bbv\right)\rho + \\\nonumber
    &\hspace{3cm}\left(\bbC_t\bbM^{-1}\bbv\right)^\top \cdot \left(\tau^{-1}\nabla U(\bbx)\right)\rho \\\nonumber
    &= \tau^{-1}\big[-\nabla U(\bbx)^\top \bbC_t^\top \cdot \left(\bbM^{-1}\bbv\right) + \\\nonumber
    &\hspace{3cm} \bbv^\top(\bbM^{-1})^\top\bbC_t^\top \cdot \left(\nabla U(\bbx)\right)\big]\rho.   
\end{align}
Given that $\bbC_t$ is symmetric, we get

\begin{equation}
\label{eq:liouville_op}
    -\nabla U(\bbx)^\top \bbC_t^\top \cdot \left(\bbM^{-1}\bbv\right) + \bbv^\top(\bbM^{-1})^\top\bbC_t^\top \cdot \left(\nabla U(\bbx)\right) = 0
\end{equation}  

Now focus on the second term in~\eqref{eq:FP_ULD}:
\begin{align}
    \label{eq:o-h_process}
    \nabla_\bbv \cdot \left(\gamma\bbv \rho \right) + &\frac{1}{2} \nabla_\bbv \cdot \nabla_\bbv \left(2\gamma \tau \bbM \rho\right)\\\nonumber
    &= \gamma\nabla_\bbv \cdot \left(\bbv \rho \right) - \gamma\nabla_\bbv \cdot \left(\tau \bbM \bbM^{-1} \tau^{-1}\bbv \rho\right) \\\nonumber
    &= \gamma\nabla_\bbv \cdot \left(\bbv \rho \right) - \gamma\nabla_\bbv \cdot \left(\bbv \rho\right) \\\nonumber
    &= 0.
\end{align}
Combining~\eqref{eq:liouville_op} and~\eqref{eq:o-h_process}, we see that $\rho(\bbx,\bbv,t) = \pi(\bbx, \bbv)$ is in the kernel of the Fokker-Planck equation, completing the proof.

\section{Proof of Proposition~\ref{prop2}}\label{A2}

The proof is similar to that of Proposition~\ref{prop1}.
We define the following matrices
\begin{equation}
    \bbD = \begin{bmatrix}
        0 & 0 & 0\\
        0 & 0 & 0\\
        0 & 0 & \alpha\bbM
    \end{bmatrix},
    \hspace{0.5cm}
    \bbQ_t = \begin{bmatrix}
        0 & -\bbC_t & 0 \\
        \bbC_t & 0 & -\lambda \bbM \\
        0 & \lambda \bbM & 0
    \end{bmatrix},
\end{equation}
and follow the same procedure as Appendix~\ref{A1}.

We need to show that the Hamiltonian of the pre-conditioned third-order dynamic is given by $H(\bbx, \bbv, \bbz) = U(\bbx) + \frac{1}{2} \bbv^\top \bbM^{-1}\bbv + \frac{1}{2} \bbz^\top \bbM^{-1}\bbz$.
Therefore, we start describing the Fokker-Planck equation~\eqref{eq:fokker_plank}  of~\eqref{eq:ct_underdamped_langevin_precond}.
The drift vector is given by
\begin{equation}
    \bbb(\bbx,\bbv, \bbz, t) = 
    \begin{bmatrix}
         \bbC_t\bbM^{-1}\bbv \\
         -\bbC_t\nabla U(\bbx) + \lambda\bbz \\
         -\lambda \bbv - \alpha \bbz
    \end{bmatrix}
\end{equation}
and the diffusion matrix is given by
\begin{equation}
    \boldsymbol{\Sigma}(\bbx,\bbv, \bbz, t)
    = 
    \begin{bmatrix}
        \bb0 \\
        \bb0 \\
        2 \tau \alpha \bbM
    \end{bmatrix}.
\end{equation}
Thus, we have
\begin{align}\label{eq:FP_GLD}
    \frac{\partial \rho}{\partial t} &= \left[\nabla_\bbx, \nabla_\bbv, \nabla_\bbz \right] \cdot \Bigg(-    
    \begin{bmatrix}
         \bbC_t\bbM^{-1}\bbv\\
        -\bbC_t\nabla U(\bbx) + \lambda\bbz\\
        -\lambda \bbv - \alpha \bbz
    \end{bmatrix}\rho + \\\nonumber 
    &\hspace{5cm}\frac{1}{2} \nabla \cdot \left( \begin{bmatrix}
        \bb0 \\
        \bb0 \\
        2\tau \alpha \bbM
    \end{bmatrix}\rho\right)\Bigg) \\
    &= \underbrace{\nabla_\bbv \cdot \left(\bbC_t\nabla U(\bbx)\rho\right) - \nabla_\bbx\cdot \left(\bbC_t\bbM^{-1}\bbv \rho\right)}_{\text{Liouville operator of Hamiltonian dynamics}}\\
    \nonumber 
    & \hspace{1.5cm}- \nabla_\bbv \cdot (\lambda \bbz\rho) + \nabla_\bbz \cdot (\lambda \bbv\rho)\\
    \nonumber
    &\hspace{1.5cm}+\underbrace{\nabla_\bbz \cdot \left(\alpha \bbz\rho\right) + \frac{1}{2} \nabla_\bbz \cdot \nabla_\bbz
        \left(2\tau \alpha \bbM \rho\right)}_{\ccalF_0 \text{ of the Ornstein–Uhlenbeck process}}
\end{align}
Both terms associated with the Liouville operator and the Ornstein–Uhlenbeck process are the same as in Appendix~\ref{A1}, so they are 0.
Then, we have
\begin{align}
    -\nabla_\bbv (\lambda \bbz \rho) + \nabla_\bbz (\lambda \bbv \rho) &= \\\nonumber
    & \hspace{-2cm} -\lambda[\bbz^\top \cdot (\tau^{-1} \bbM^{-1}\bbv)]\rho + \lambda[\bbv^\top \cdot (\tau^{-1} \bbM^{-1}\bbz)]\rho.
\end{align}
This is $0$ due to the symmetry of $\bbM$, completing the proof.

\end{appendices}

\bibliographystyle{IEEEbib}
\bibliography{citations}

\end{document}


\markboth{IEEE TRANSACTIONS ON SIGNAL PROCESSING (SUBMITTED)}%
         {Zilberstein et al. \MakeLowercase{\textit{et al.}}: \theTitle}

\title{Supplementary material - Solving Linear Inverse Problems using Higher-Order Annealed Langevin Diffusion\\
\thanks{}
}

\author{%
      \IEEEauthorblockN{Nicolas Zilberstein, Ashutosh Sabharwal, Santiago Segarra}  
}

\maketitle

\section{Images of experiments in Section V-C.}

\begin{figure}[H]
    \includegraphics[width=.135\textwidth]{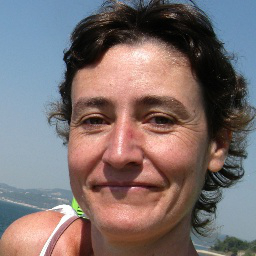}
    \includegraphics[width=.135\textwidth]{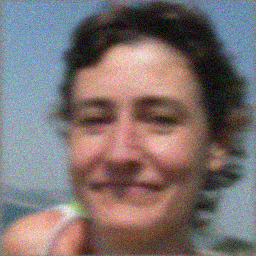}
    \includegraphics[width=.135\textwidth]{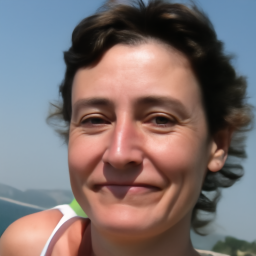}
\includegraphics[width=.135\textwidth]{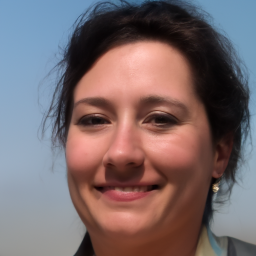}
    \includegraphics[width=.135\textwidth]{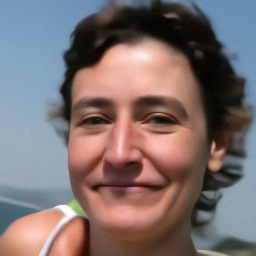}
    \includegraphics[width=.135\textwidth]{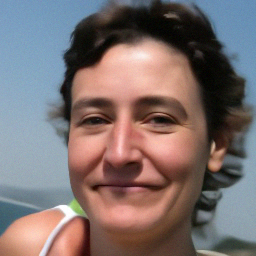}  \includegraphics[width=.135\textwidth]{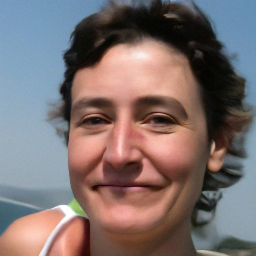}  \\[\smallskipamount]
    \includegraphics[width=.135\textwidth]{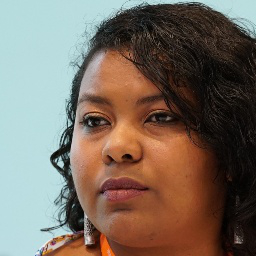}
    \includegraphics[width=.135\textwidth]{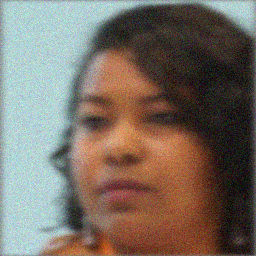}
    \includegraphics[width=.135\textwidth]{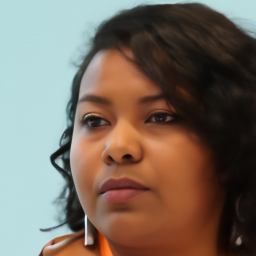}
\includegraphics[width=.135\textwidth]{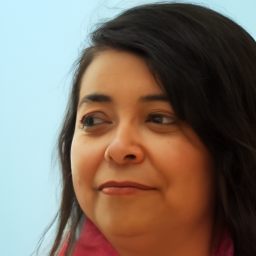}
    \includegraphics[width=.135\textwidth]{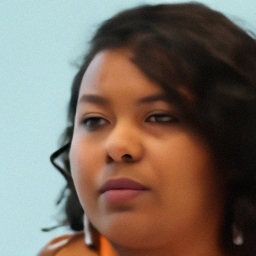}    \includegraphics[width=.135\textwidth]{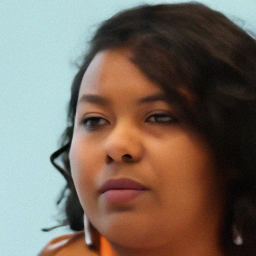}  \includegraphics[width=.135\textwidth]{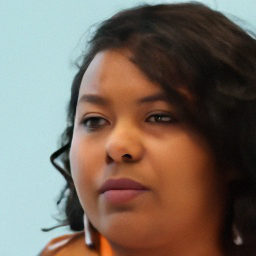} \\[\smallskipamount]
    \includegraphics[width=.135\textwidth]{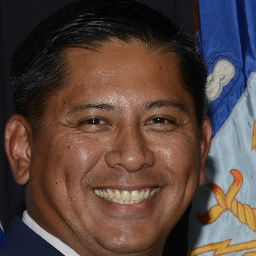}
    \includegraphics[width=.135\textwidth]{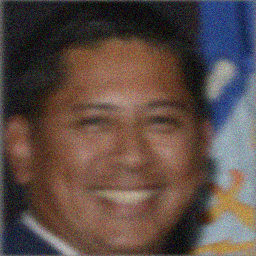}
    \includegraphics[width=.135\textwidth]{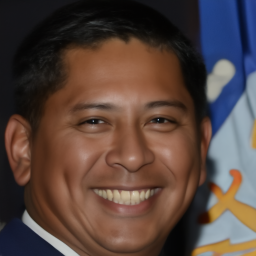}
\includegraphics[width=.135\textwidth]{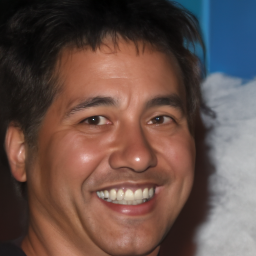}
    \includegraphics[width=.135\textwidth]{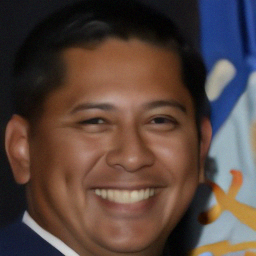}    \includegraphics[width=.135\textwidth]{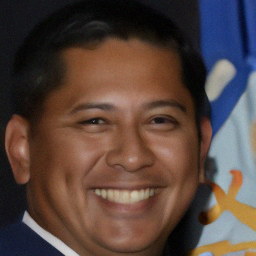}  \includegraphics[width=.135\textwidth]{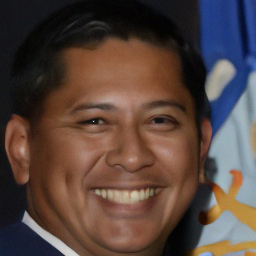} \\[\smallskipamount]
    \includegraphics[width=.135\textwidth]{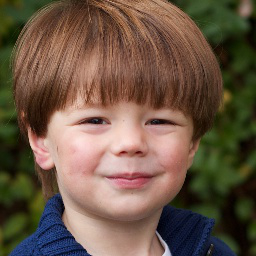}
    \includegraphics[width=.135\textwidth]{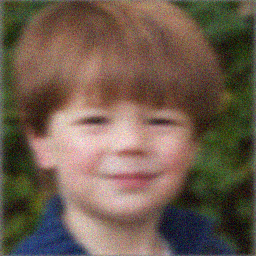}    \includegraphics[width=.135\textwidth]{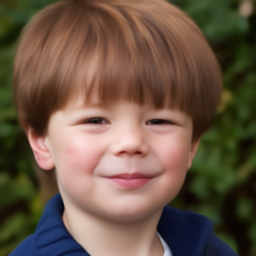}
    \includegraphics[width=.135\textwidth]{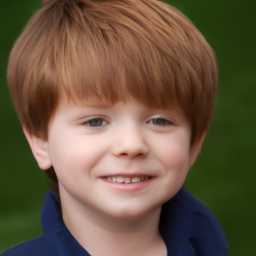}
    \includegraphics[width=.135\textwidth]{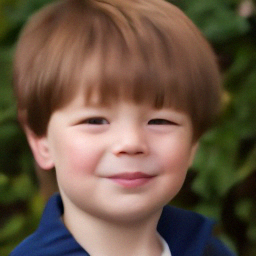}    \includegraphics[width=.135\textwidth]{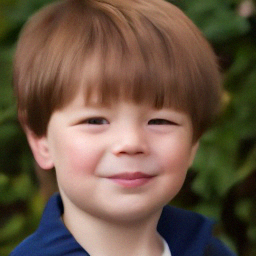}
    \includegraphics[width=.135\textwidth]{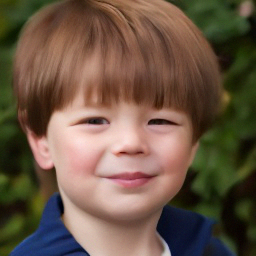}
    \\[\smallskipamount]
    \caption{Gaussian deblurring with anisotropic filter on FFHQ with $\sigma_0 = 0.05$. From left to right:
    true image, measurement, DDRM, DPS, SNIPS, Underd. Lang., 3rd order Lang.}\label{fig:foobar}
\end{figure}

\begin{figure}[H]
    \includegraphics[width=.135\textwidth]{figures/original_deblur/orig_0.png}
    \includegraphics[width=.135\textwidth]{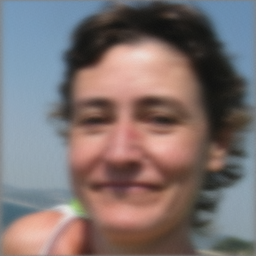}        \includegraphics[width=.135\textwidth]{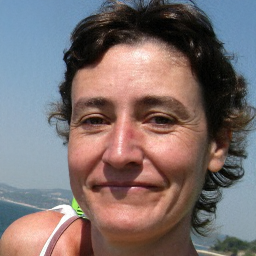}
    \includegraphics[width=.135\textwidth]{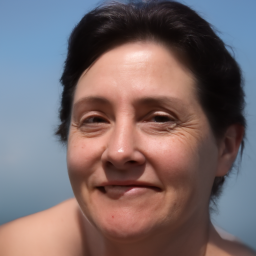} 
    \includegraphics[width=.135\textwidth]{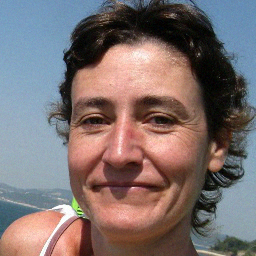} 
    \includegraphics[width=.135\textwidth]{figures/recon_deblur_noisy/recon_0_2ndorder.png}      \includegraphics[width=.135\textwidth]{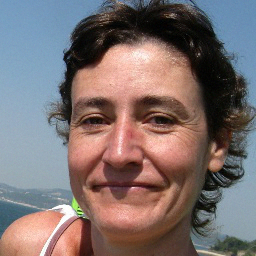}  \\[\smallskipamount]
    \includegraphics[width=.135\textwidth]{figures/original_deblur/orig_1.png}
    \includegraphics[width=.135\textwidth]{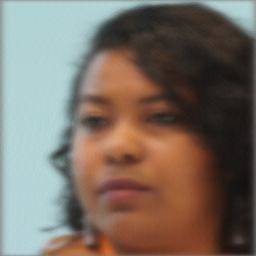}        \includegraphics[width=.135\textwidth]{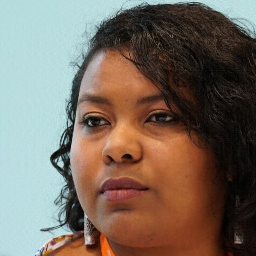}
    \includegraphics[width=.135\textwidth]{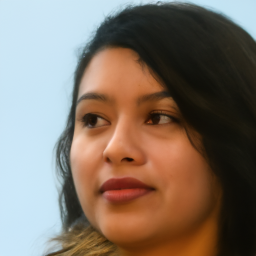} 
    \includegraphics[width=.135\textwidth]{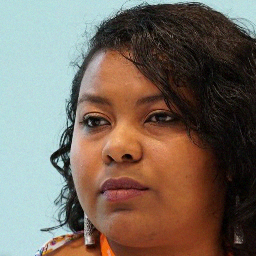}    \includegraphics[width=.135\textwidth]{figures/recon_deblur_noisy/recon_1_2ndorder.png}      \includegraphics[width=.135\textwidth]{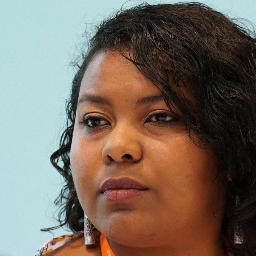}  \\[\smallskipamount]
    \includegraphics[width=.135\textwidth]{figures/original_deblur/orig_2.png}
     \includegraphics[width=.135\textwidth]{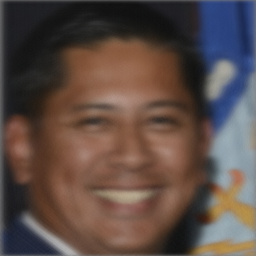}        \includegraphics[width=.135\textwidth]{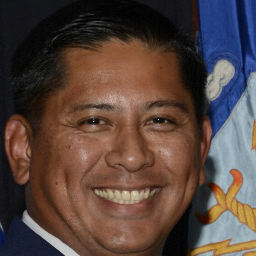}
    \includegraphics[width=.135\textwidth]{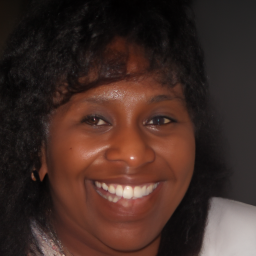} 
    \includegraphics[width=.135\textwidth]{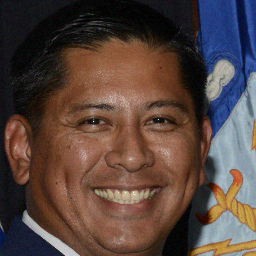}    \includegraphics[width=.135\textwidth]{figures/recon_deblur_noisy/recon_2_2ndorder.png}      \includegraphics[width=.135\textwidth]{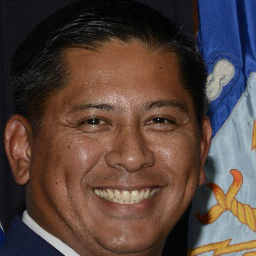} \\[\smallskipamount]
    \includegraphics[width=.135\textwidth]{figures/original_deblur/orig_3.png}
    \includegraphics[width=.135\textwidth]{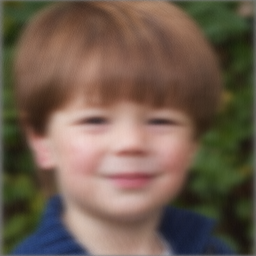}    
    \includegraphics[width=.135\textwidth]{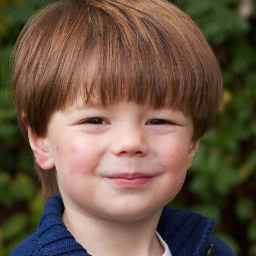}
    \includegraphics[width=.135\textwidth]{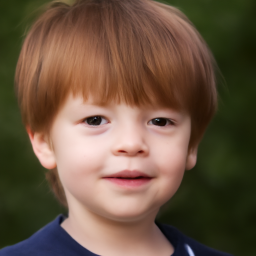} 
    \includegraphics[width=.135\textwidth]{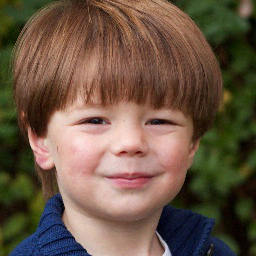}    \includegraphics[width=.135\textwidth]{figures/recon_deblur_noisy/recon_3_2ndorder.png}
    \includegraphics[width=.135\textwidth]{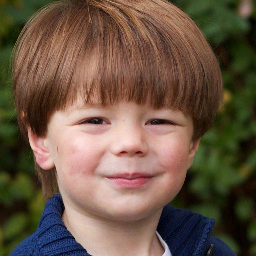}   
    \\[\smallskipamount]
    \caption{Gaussian deblurring with anisotropic filter on FFHQ with $\sigma_0 = 0.0002$. From left to right:
    true image, measurement, DDRM, DPS, SNIPS, Underd. Lang., 3rd order Lang.}\label{fig:foobar}
\end{figure}

\begin{figure}[H]
    \includegraphics[width=.135\textwidth]{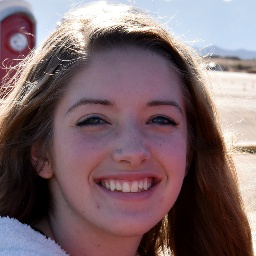}
    \includegraphics[width=.135\textwidth]{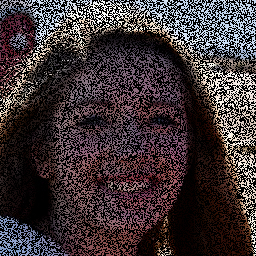}        \includegraphics[width=.135\textwidth]{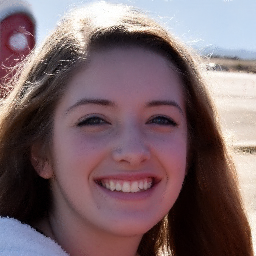}
    \includegraphics[width=.135\textwidth]{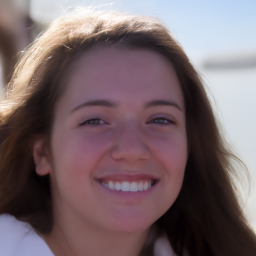} 
    \includegraphics[width=.135\textwidth]{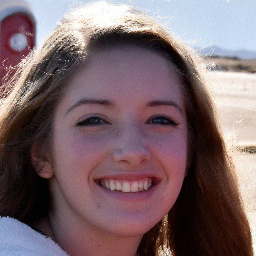}   
\includegraphics[width=.135\textwidth]{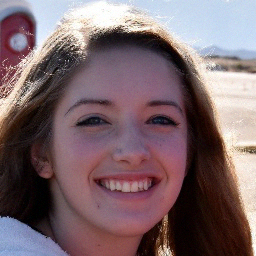}   \includegraphics[width=.135\textwidth]{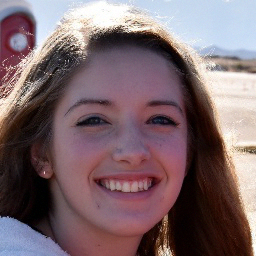}\\[\smallskipamount]
    \includegraphics[width=.135\textwidth]{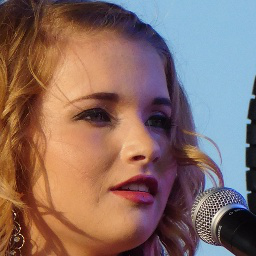}
    \includegraphics[width=.135\textwidth]{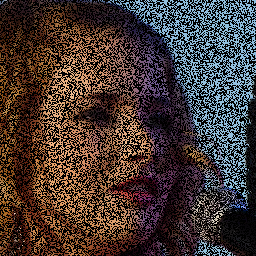}         \includegraphics[width=.135\textwidth]{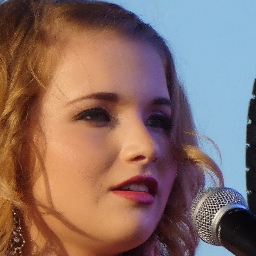}
    \includegraphics[width=.135\textwidth]{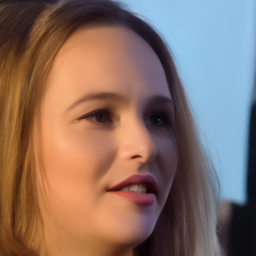} 
    \includegraphics[width=.135\textwidth]{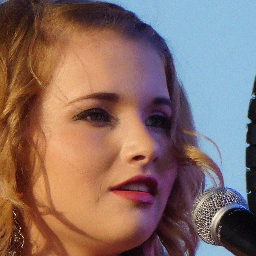}    \includegraphics[width=.135\textwidth]{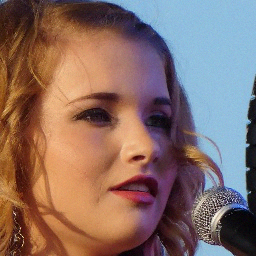}     \includegraphics[width=.135\textwidth]{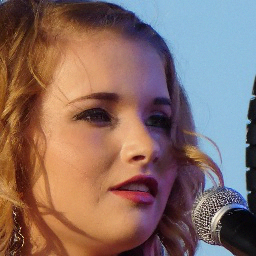}\\[\smallskipamount]
    \includegraphics[width=.135\textwidth]{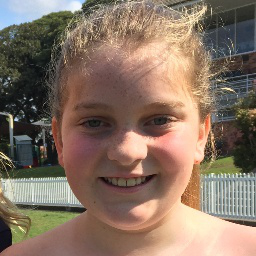}
    \includegraphics[width=.135\textwidth]{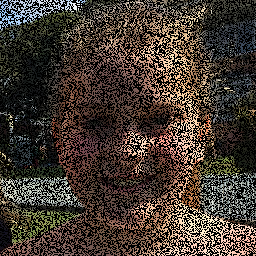}         \includegraphics[width=.135\textwidth]{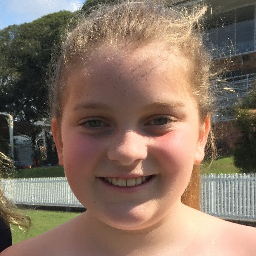}
    \includegraphics[width=.135\textwidth]{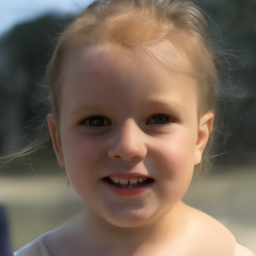} 
    \includegraphics[width=.135\textwidth]{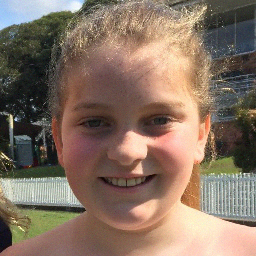}     \includegraphics[width=.135\textwidth]{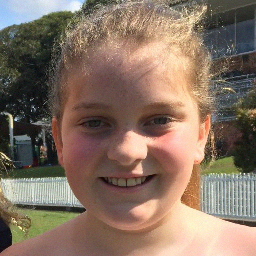}     \includegraphics[width=.135\textwidth]{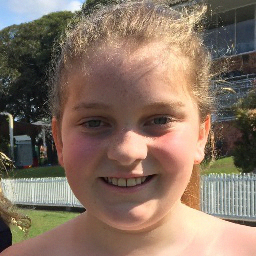}\\[\smallskipamount]
    \includegraphics[width=.135\textwidth]{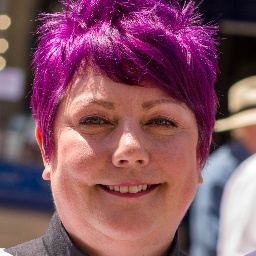}
    \includegraphics[width=.135\textwidth]{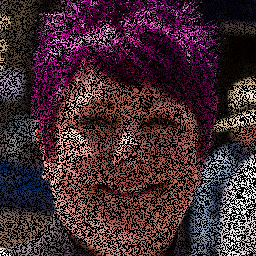}    
    \includegraphics[width=.135\textwidth]{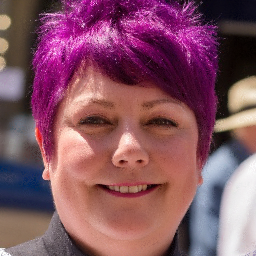}
    \includegraphics[width=.135\textwidth]{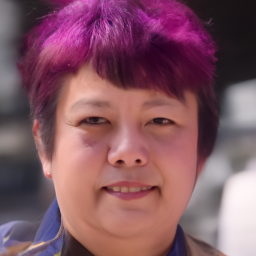} 
    \includegraphics[width=.135\textwidth]{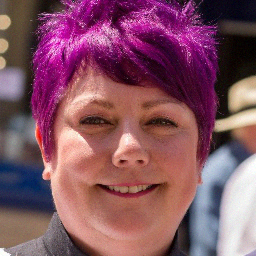}    \includegraphics[width=.135\textwidth]{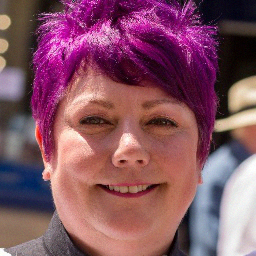}
    \includegraphics[width=.135\textwidth]{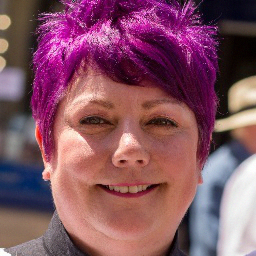}   
    \\[\smallskipamount]
    \caption{Inpainting on FFHQ with $\sigma_0 = 0.0002$; we randomly drop 50\% of the pixels. From left to right:
    true image, measurement, DDRM, DPS, SNIPS, Underd. Lang., 3rd order Lang.}\label{fig:foobar}
\end{figure}

\begin{figure}[H]
    \includegraphics[width=.19\textwidth]{figures/original_inp/orig_6.png}
    \includegraphics[width=.19\textwidth]{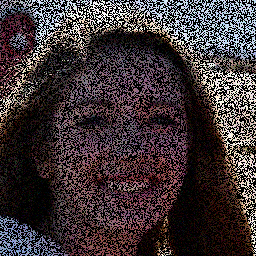}         
    \includegraphics[width=.19\textwidth]{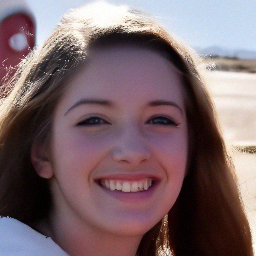} 
    \includegraphics[width=.19\textwidth]{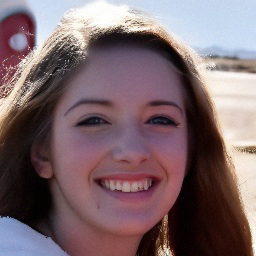}   \includegraphics[width=.19\textwidth]{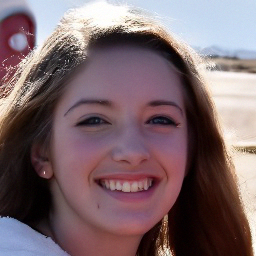}  \\[\smallskipamount]
    \includegraphics[width=.19\textwidth]{figures/original_inp/orig_7.png}
    \includegraphics[width=.19\textwidth]{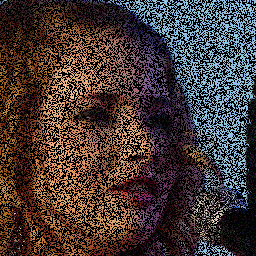}         
    \includegraphics[width=.19\textwidth]{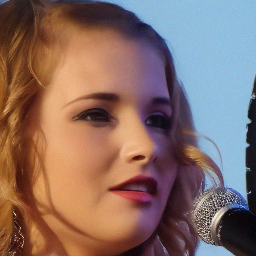}   \includegraphics[width=.19\textwidth]{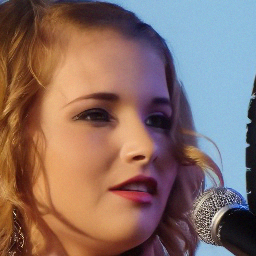}   \includegraphics[width=.19\textwidth]{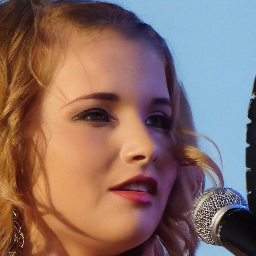}  \\[\smallskipamount]
    \includegraphics[width=.19\textwidth]{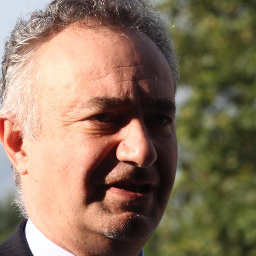}
    \includegraphics[width=.19\textwidth]{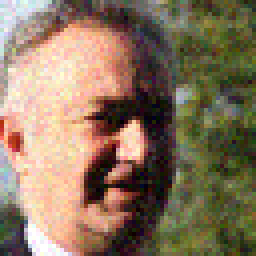}   \includegraphics[width=.19\textwidth]{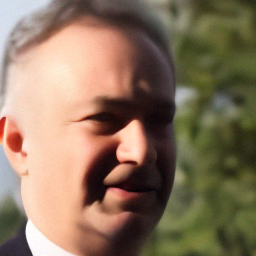}     \includegraphics[width=.19\textwidth]{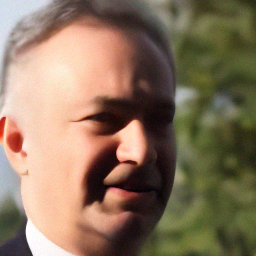}      \includegraphics[width=.19\textwidth]{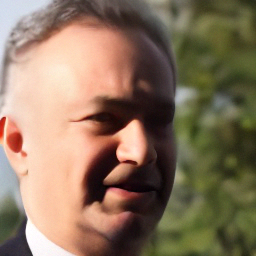}  \\[\smallskipamount]
    \includegraphics[width=.19\textwidth]{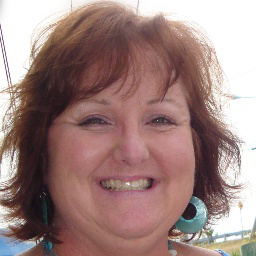}
    \includegraphics[width=.19\textwidth]{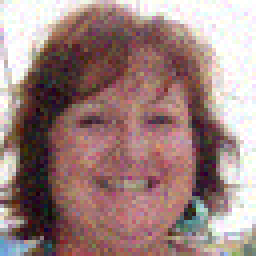}   
    \includegraphics[width=.19\textwidth]{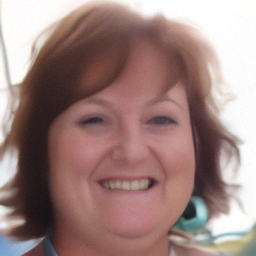}        \includegraphics[width=.19\textwidth]{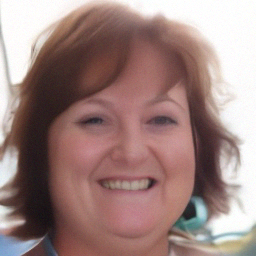}   
    \includegraphics[width=.19\textwidth]{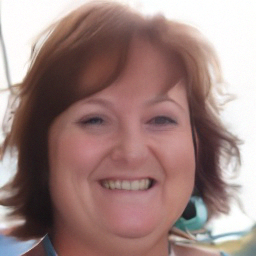}   
    \\[\smallskipamount]
    \caption{Inpainting (first two rows) and super-resolution $\times 4$ (last two rows) on FFHQ with $\sigma_0 = 0.05$.  From left to right: true image, measurement, SNIPS, Underd. Lang., 3rd order Lang.}\label{fig:foobar}
\end{figure}

\section{Additional experiments}

We compute additional quantitative results based on the standard SSIM metric in Table~\ref{table:results_noiseless_images_SSIM} and Table~\ref{table:results_noisy_images_SSIM}.
The results for this metric are similar than for the other ones: for the noiseless case, the 3rd order Lang works better than the other methods, while for the noisy case DDRM outperforms the other methods for deblurring and super-resolution.

\begin{table}[H]
    \centering
    \caption{{\small SSIM on noiseless images ($\sigma_0 = 0.0002$) from FFHQ. The best method for each metric and experiment is bolded.}}
    \begin{tabular}{c | c|c | c}
         & \multicolumn{1}{c }{Deblurr (Gauss)} & \multicolumn{1}{c }{Inpainting} & \multicolumn{1}{c }{SRx4} \\ 
        \hline
        Method & SSIM $\uparrow$ & SSIM $\uparrow$ &  SSIM $\uparrow$\\ 
        \hline 
        DDRM & 0.95 & 0.9 & 0.83\\ 
        DPS & 0.51 & 0.53  & 0.41 \\ 
        SNIPS & 0.93 & 0.89 & 0.83 \\ 
        Underd. Lang & 0.94 & 0.9  & 0.83 \\ 
        3rd. order Lang & \textbf{0.95} & \textbf{0.92} & \textbf{0.84}\\ 
    \end{tabular}
    \label{table:results_noiseless_images_SSIM}
\end{table}

\begin{table}[H]
    \centering
    \caption{{\small SSIM on noisy images ($\sigma_0 = 0.05$) from FFHQ. The best method for each metric and experiment is bolded.}}
    \begin{tabular}{c | c|c | c}
         & \multicolumn{1}{c }{Deblurr (Gauss)} & \multicolumn{1}{c }{Inpainting} & \multicolumn{1}{c }{SRx4} \\ 
        \hline
        Method & SSIM $\uparrow$ & SSIM $\uparrow$ &  SSIM $\uparrow$\\ 
        \hline 
        DDRM & \textbf{0.81} & 0.84 & \textbf{0.78}\\ 
        DPS & 0.51 & 0.52 & 0.41\\ 
        SNIPS & 0.77  & 0.84 &  0.75\\ 
        Underd. Lang & 0.76 & 0.83 & 0.74 \\ 
        3rd. order Lang & 0.77 & \textbf{0.85} & 0.76\\ 
    \end{tabular}
    \label{table:results_noisy_images_SSIM}
\end{table}